%% file: ms.tex
\documentclass[sigconf]{acmart}

\usepackage[utf8]{inputenc} 
\usepackage[T1]{fontenc}    
\usepackage{hyperref}       
\usepackage{url}            
\usepackage{booktabs}       
\usepackage{amsfonts}       
\usepackage{nicefrac}       
\usepackage{microtype}      
\usepackage{enumitem}       
\usepackage{algorithm}
\usepackage{xspace}
\usepackage{multirow}
\usepackage{dsfont}
\usepackage{amsmath} 
\usepackage{graphicx}
\usepackage{subcaption}
\usepackage[noend]{algpseudocode}
\usepackage{olo}

\newcommand{\eg}{{\it e.g\xspace}}
\newcommand{\etc}{{\it etc\xspace}}

\newcommand{\unif}{\textrm{Unif}}
\definecolor{silver}{rgb}{0.75, 0.75, 0.75}
\newcommand{\wpred}[1]{
\textcolor{silver}{#1}\xspace
}

\setcopyright{iw3c2w3}
\copyrightyear{2021}
\acmYear{2021}
\acmDOI{10.1145/3442381.3449815}

\acmConference[WWW '21]{Proceedings of the Web Conference 2021}{April 19--23, 2021}{Ljubljana, Slovenia}
\acmBooktitle{Proceedings of the Web Conference 2021 (WWW '21), April 19--23, 2021,
Ljubljana, Slovenia}
\acmPrice{}
\acmISBN{978-1-4503-8312-7/21/04}

\def\alg{\textsf{ECLARE}\xspace}
\def\algfull{Extreme Classification with Label Graph Correlations}
\def\lt{\textsf{LTE}\xspace}
\def\ltaug{\textsf{GALE}\xspace}
\def\metaprop{\textsf{GAME}\xspace}
\def\ova{one-vs-all\xspace}
\def\Ova{One-vs-all\xspace}

\def\notsametext{\textsf{``Dog of Osu''}\xspace}
\def\testdocument{\textsf{``Tibetan Terrier''}\xspace}
\def\co{\textsf{``Tibetan Spaniel'', ``Tibetan kyi apso'' and ``Tibetan Mastiff''}\xspace}

\usepackage{pifont}

\newcommand{\code}{\href{https://github.com/Extreme-classification/ECLARE}{\color{blue}{https://github.com/Extreme-classification/ECLARE}}}\xspace

\begin{document}

\input{sections/frontmatter}
\input{sections/introduction}
\input{sections/literature_survey}
\input{sections/method}
\input{sections/experiments}
\input{sections/conclusion}

\begin{acks}
The authors thank the reviewers for helpful comments that improved the presentation of the paper. They also thank the IIT Delhi HPC facility for computational resources. AM is supported by a Google PhD Fellowship.
\end{acks}

\bibliographystyle{ACM-Reference-Format}
\bibliography{ms}

\end{document}

%% file: sections/frontmatter.tex
\title{\alg: \algfull}

\author{Anshul Mittal}
\email{me@anshulmittal.org}
\affiliation{%
    \institution{IIT Delhi}
    \country{India}
}
\author{Noveen Sachdeva}
\authornote{Work done during an internship at Microsoft Research India}
\email{nosachde@ucsd.edu}
\affiliation{%
    \institution{UC San Diego}
    \country{USA}
}
\author{Sheshansh Agrawal}
\email{sheshansh.agrawal@microsoft.com}
\affiliation{%
    \institution{Microsoft Research}
    \country{India}
}
\author{Sumeet Agarwal}
\orcid{0000-0002-5714-3921}
\email{sumeet@iitd.ac.in}
\affiliation{%
    \institution{IIT Delhi}
    \country{India}
}
\author{Purushottam Kar}
\orcid{0000-0003-2096-5267}
\email{purushot@cse.iitk.ac.in}
\affiliation{%
    \institution{IIT Kanpur}
    \institution{Microsoft Research}
    \country{India}
}
\author{Manik Varma}
\email{manik@microsoft.com}
\affiliation{%
    \institution{Microsoft Research}
    \institution{IIT Delhi}
    \country{India}
}

\renewcommand{\shortauthors}{Mittal and Sachdeva, et al.}

\begin{abstract}
Deep extreme classification (XC) seeks to train deep architectures that can tag a data point with its most relevant subset of labels from an extremely large label set. The core utility of XC comes from predicting labels that are rarely seen during training. Such \emph{rare} labels hold the key to personalized recommendations that can delight and surprise a user. However, the large number of rare labels and small amount of training data per rare label offer significant statistical and computational challenges. State-of-the-art deep XC methods attempt to remedy this by incorporating textual descriptions of labels but do not adequately address the problem. This paper presents \alg, a scalable deep learning architecture that incorporates not only label text, but also label correlations, to offer accurate real-time predictions within a few milliseconds. Core contributions of \alg include a frugal architecture and scalable techniques to train deep models along with label correlation graphs at the scale of millions of labels. In particular, \alg offers predictions that are 2--14\% more accurate on both publicly available benchmark datasets as well as proprietary datasets for a related products recommendation task sourced from the Bing search engine. Code for \alg is available at \code
\end{abstract}

\ccsdesc[500]{Computing methodologies~Machine learning}
\ccsdesc[300]{Computing methodologies~Supervised learning by classification}

\keywords{Extreme multi-label classification; product to product recommendation; label features; label metadata; large-scale learning}

\maketitle

%% file: sections/introduction.tex
\section{Introduction}
\label{sec:intro}

\noindent\textbf{Overview.} Extreme multi-label classification (XC) involves tagging a data point with the subset of labels most relevant to it, from an extremely large set of labels. XC finds applications in several domains including product recommendation \cite{Medini2019}, related searches \cite{Jain19}, related products \cite{mittal20}, \etc. This paper demonstrates that XC methods stand to benefit significantly from utilizing label correlation data, by presenting \alg, an XC method that utilizes textual label descriptions and label correlation graphs over millions of labels to offer predictions that can be 2--14\% more accurate than those offered by state-of-the-art XC methods, including those that utilize label metadata such as label text.\\

\input{figures/fig_tex/labal_pop}

\noindent\textbf{Rare Labels.} XC applications with millions of labels typically find that most labels are \emph{rare}, with very few training data points tagged with those labels. Fig~\ref{fig:lblpop} exemplifies this on two benchmark datasets where 60--80\% labels have $< 5$ training points. The reasons behind rare labels are manifold. In several XC applications, there may exist an inherent skew in the popularity of labels, \eg, it is natural for certain products to be more popular among users on an e-commerce platform. XC applications also face \emph{missing labels} \cite{Jain16,Yu14} where training points are not tagged with all the labels relevant to them. Reasons for this include the inability of human users to exhaustively mark all products of interest to them, and biases in the recommendation platform (\eg. website, app) itself which may present or impress upon its users, certain products more often than others.

\noindent\textbf{Need for Label Metadata.} Rare labels are of critical importance in XC applications. They allow highly personalized yet relevant recommendations that may delight and surprise a user, or else allow precise and descriptive tags to be assigned to a document, \etc. However, the paucity of training data for rare labels makes it challenging to predict them accurately. Incorporating label metadata such as textual label descriptions \cite{mittal20}, label taxonomies \cite{Kanagal12, menon11, sachdeva19} and label co-occurrence into the classification process are possible ways to augment the available information for rare labels.

\noindent\textbf{Key Contributions of \alg.} This paper presents \alg, an XC method that performs collaborative learning that benefits rare labels. This is done by incorporating multiple forms of label metadata such as label text as well as dynamically inferred label correlation graphs. Critical to \alg are augmentations to the architecture and learning pipeline that scale to millions of labels:
\begin{enumerate}[leftmargin=*]
    \item Introduce a framework that allows collaborative extreme learning using label-label correlation graphs that are dynamically generated using asymmetric random walks. This is in contrast to existing approaches that often perform collaborative learning on static user-user or document-document graphs~\cite{He2020,hamilton17,ying2018}.
    \item Introduce the use of multiple representations for each label: one learnt from label text alone (\lt), one learnt collaboratively from label correlation graphs (\ltaug), and a label-specific refinement vector. \alg proposes a robust yet inexpensive attention mechanism to fuse these multiple representations to generate a single \ova classifier per label.
    \item Propose critical augmentations to well-established XC training steps, such as label clustering, negative sampling, classifier initialization,  shortlist creation (\metaprop), \etc, in order to incorporate label correlations in a systematic and scalable manner.
    \item Offer an end-to-end training pipeline incorporating the above components in an efficient manner which can be scaled to tasks with millions of labels and offer up to 14\% performance boost on standard XC prediction metrics.
\end{enumerate}

\noindent\textbf{Comparison to State-of-the-art.} Experiments indicate that apart from significant boosts on standard XC metrics (see Tab~\ref{tab:baelines_eval_p2p}), \alg offers two distinct advantages over existing XC algorithms, including those that do use label metadata such as label text (see Tab~\ref{tab:examples})
\begin{enumerate}[leftmargin=*]
    \item \textbf{Superior Rare Label Prediction}: In the first example in Tab~\ref{tab:examples}, for the document \testdocument, \alg correctly predicts the rare label \notsametext that appeared just twice in the training set. All other methods failed to predict this rare label. It is notable that this label has no common tokens (words) with the document text or other labels which indicates that relying solely on label text is insufficient. \alg offers far superior performance on \emph{propensity scored} XC metrics which place more emphasis on predicting rare labels correctly (see Tabs~\ref{tab:baelines_eval_p2p} and \ref{tab:p2p}).
    \item \textbf{Superior Intent Disambiguation}: The second and third examples in Tab~\ref{tab:examples} further illustrate pitfalls of relying on label text alone as metadata. For the document ``\textsf{85th Academy Awards}'', all other methods are incapable of predicting other award ceremonies held in the same year and make poor predictions. On the other hand, \alg was better than other methods at picking up subtle cues and associations present in the training data to correctly identify associated articles. \alg offers higher precision@1 and recall@10 (see Tabs~\ref{tab:baelines_eval_p2p} and \ref{tab:p2p}).
\end{enumerate}

%% file: figures/fig_tex/labal_pop.tex
\begin{figure}
    \centering
    \begin{subfigure}{0.4\linewidth}
      \centering
      \includegraphics[width=\linewidth]{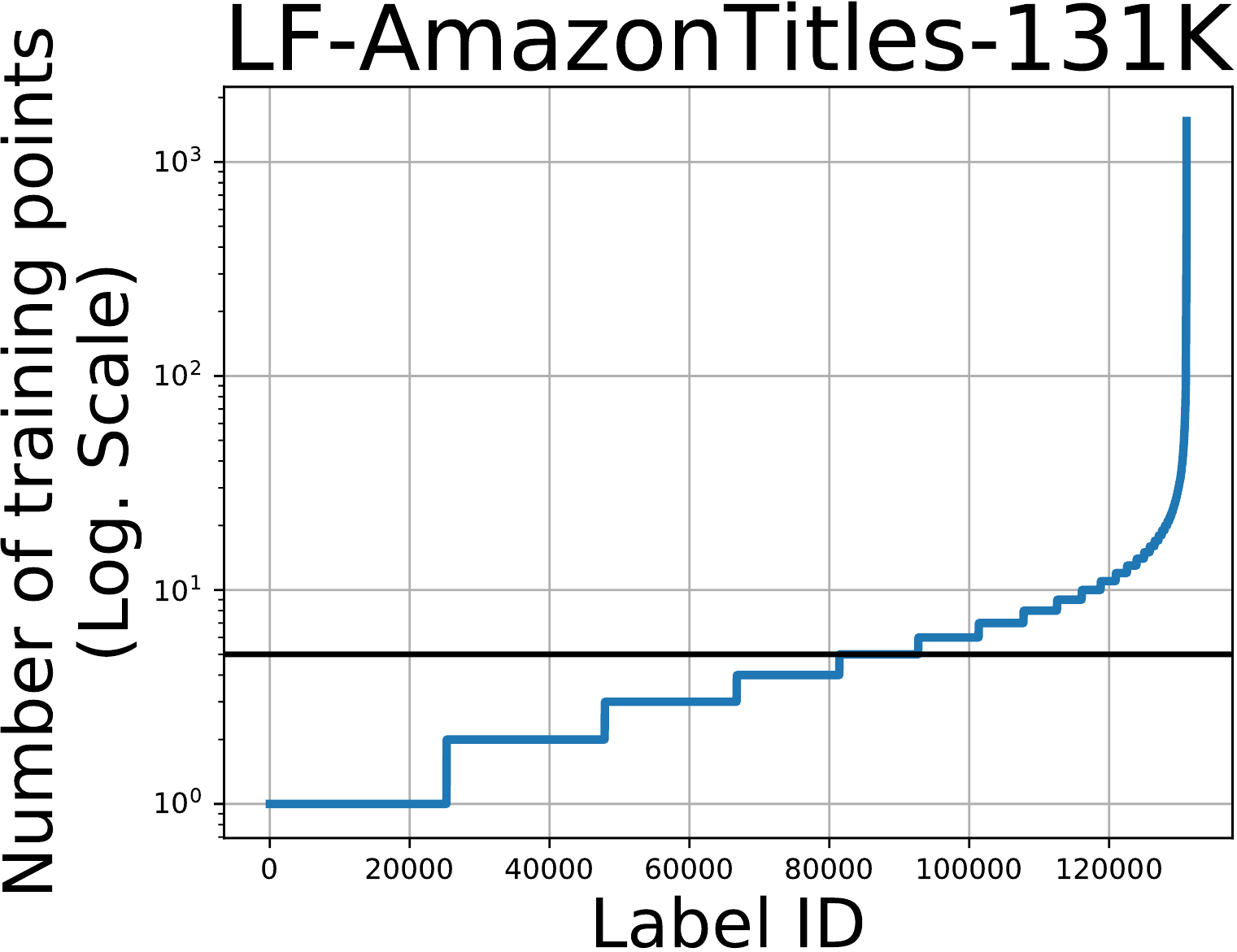}
    \end{subfigure}~
    \begin{subfigure}{0.45\linewidth}
      \centering
        \includegraphics[width=\linewidth]{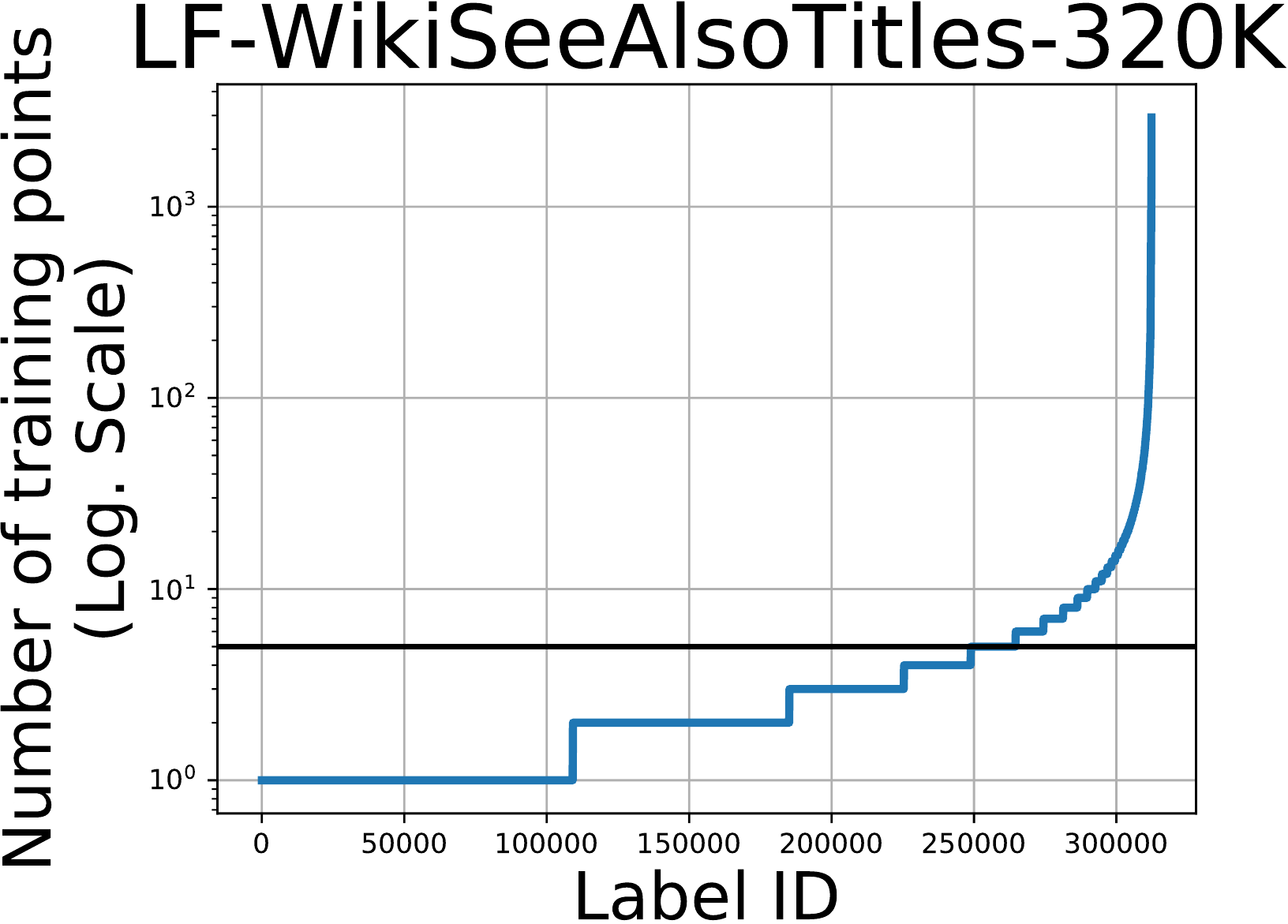}
    \end{subfigure}
    \caption{Number of training points per label for two datasets. Labels are ordered from left to right in increasing order of popularity. 60--80\% labels have $< 5$ training points (the bold horizontal line indicates the 5 training point level).}
    \label{fig:lblpop}
\end{figure}

%% file: sections/literature_survey.tex
\section{Related Work}
\label{sec:related}

\noindent\textbf{Summary.} XC algorithms proposed in literature employ a variety of label prediction approaches like tree, embedding, hashing and \ova-based approaches~\cite{Babbar17, Prabhu14, Prabhu18,  Prabhu18b, Jain16, Jain17, Jain19, Babbar19, Yen16, Yen17, Yen18a, Khandagale19, Jasinska16, Siblini18a, Jalan2019, Tagami17, Liu17, You18, Chang20, Wydmuch18, Guo2019, Bhatia15, Gupta19, Dahiya21, mittal20}. Earlier works learnt label classifiers using fixed representations for documents (typically bag-of-words) whereas contemporary approaches learn a document embedding architecture (typically using deep networks) jointly with the label classifiers. In order to operate with millions of labels, XC methods frequently have to rely on sub-linear time data structures for operations such as shortlisting labels, sampling hard negatives, \etc. Choices include hashing~\cite{Medini2019}, clustering~\cite{Prabhu18b, You18, Chang20}, negative sampling~\cite{Mikolov13}, \etc. Notably, most XC methods except DECAF~\cite{mittal20}, GLaS~\cite{Guo2019}, and X-Transformer~\cite{Chang20} do not incorporate any form of label metadata, instead treating labels as black-box identifiers.\\

\noindent\textbf{Fixed Representation.} Much of the early work in XC used fixed bag-of-words (BoW) features to represent documents. \Ova methods such as DiSMEC~\cite{Babbar17}, PPDSparse~\cite{Yen17}, ProXML~\cite{Babbar19} decompose the XC problem into several binary classification problems, one per label. Although these offered state-of-the-art performance until recently, they could not scale beyond a few million labels. To address this, several approaches were suggested to speed up training~\citep{Khandagale19, Jain19, Prabhu18b, Yen18a}, and prediction~\citep{Jasinska16, Niculescu17} using tree-based classifiers and negative sampling. These offered high performance as well as could scale to several millions of labels. However, these architectures were suited for fixed features and did not support jointly learning document representations. Attempts, such as \cite{Jain19}, to use pre-trained features such as FastText~\citep{Joulin17} were also not very successful if the features were trained on an entirely unrelated task.\\

\noindent\textbf{Representation Learning.} Recent works such as X-Transformer~\cite{Chang20}, ASTEC~\cite{Dahiya21}, XML-CNN~\cite{Liu17}, DECAF~\cite{mittal20} and AttentionXML~\cite{You18} propose architectures that jointly learn representations for the documents as well as label classifiers. For the most part, these methods outperform their counterparts that operate on fixed document representations which illustrates the superiority of task-specific document representations over generic pre-trained features. However, some of these methods utilize involved architectures such as attention \cite{You18,Chang20} or convolutions \cite{Liu17}. It has been observed \cite{Dahiya21,mittal20} that in addition to being more expensive to train, these architectures also suffer on XC tasks where the documents are short texts, such as user queries, or product titles.\\

\noindent\textbf{XC with Label Metadata.} Utilizing label metadata such as label text, label correlations, \etc. can be critical for accurate prediction of rare labels, especially on short-text applications where documents have textual descriptions containing only 5-10 tokens which are not very descriptive. Among existing works, GLaS~\cite{Guo2019} uses label correlations to design a regularizer that improved performance over rare labels, while X-Transformer~\cite{Chang20} and DECAF~\cite{mittal20} use label text as label metadata instead. X-Transformer utilizes label text to perform semantic label indexing (essentially a shortlisting step) along with a pre-trained-then-fine-tuned RoBERTa \cite{roberta} architecture. On the other hand, DECAF uses a simpler architecture to learn both label and document representations in an end-to-end manner.

\noindent\textbf{Collaborative Learning for XC.} Given the paucity of data for rare labels, the use of label text alone can be insufficient to ensure accurate prediction, especially in short-text applications such as related products and related queries search, where the amount of label text is also quite limited. This suggests using label correlations to perform collaborative learning on the label side. User-user \emph{or} document-document graphs \cite{ying2018, He2020, Kipf17, Tang20, zhang20, hamilton17, velickovic18, pal20} have become popular, with numerous methods such as GCN~\cite{Kipf17}, LightGCN~\cite{He2020}, GraphSAGE~\cite{hamilton17}, PinSage~\cite{ying2018}, \etc. utilizing graph neural networks to augment user/document representations. However, XC techniques that directly enable label collaboration with millions of labels have not been explored. One of the major barriers for this seems to be that label correlation graphs in XC applications turn out to be extremely sparse, \eg, for the label correlation graph \alg constructed for the LF-WikiSeeAlsoTitles-320K dataset, nearly 18\% of labels had no edges to any other label. This precludes the use of techniques such as Personalised Page Rank (PPR)~\cite{ying2018, klicpera18} over the ground-truth to generate a set of shortlisted labels for negative sampling. \alg solves this problem by first mining hard-negatives for each label using a separate technique, and subsequently augmenting this list by adding highly correlated labels.

%% file: sections/method.tex
\section{\alg: \algfull}
\label{sec:method}

\begin{figure}
    \centering
    \includegraphics[width=0.86\columnwidth]{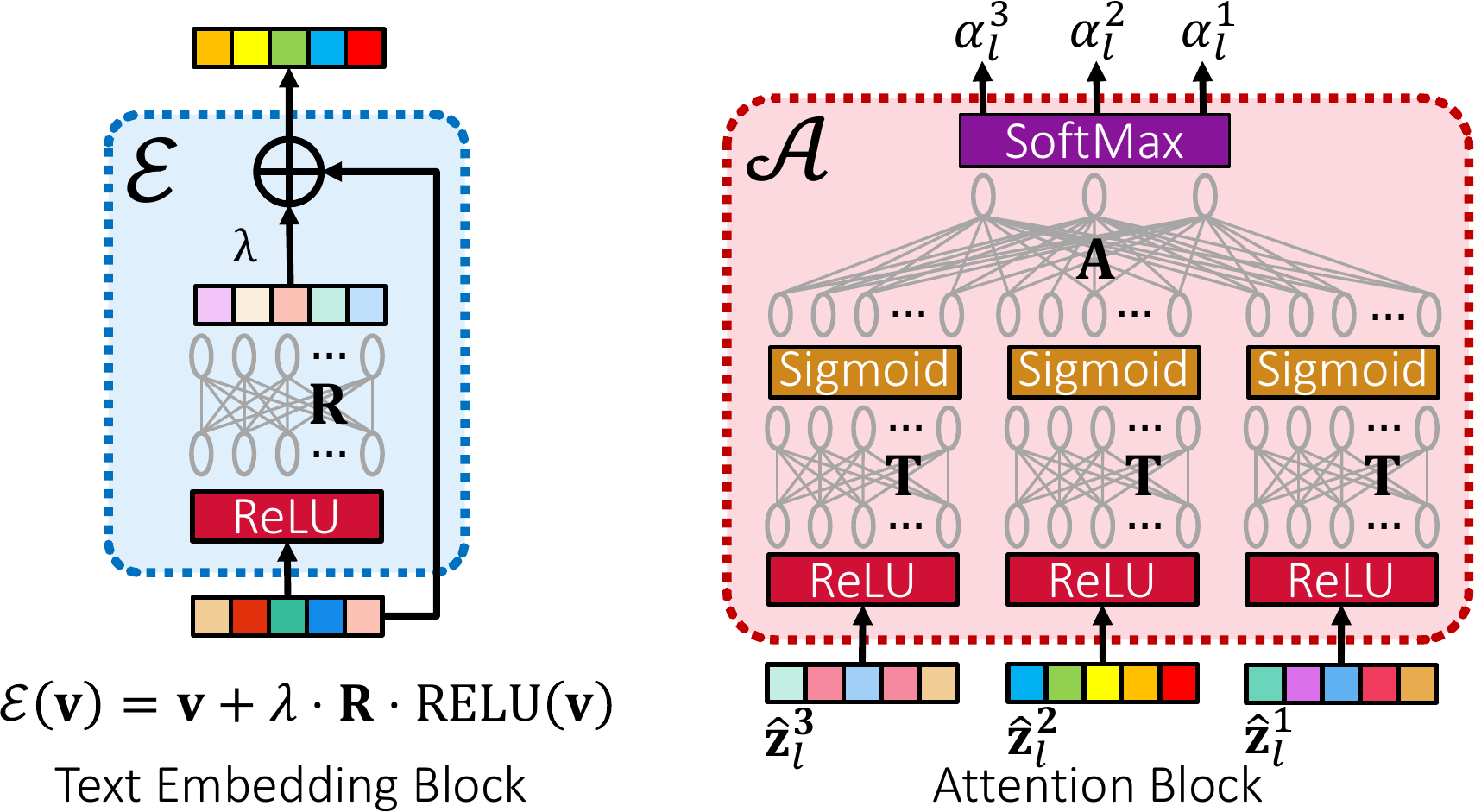}
    \caption{The building blocks of \alg. (Left) The embedding block is used in document and label embeddings. (Right) The attention block is used to fuse multiple label representations into a single label classifier (see Fig~\ref{fig:embedding}).}
    \label{fig:block}
\end{figure}

\begin{figure}
    \centering
    \includegraphics[width=0.93\columnwidth]{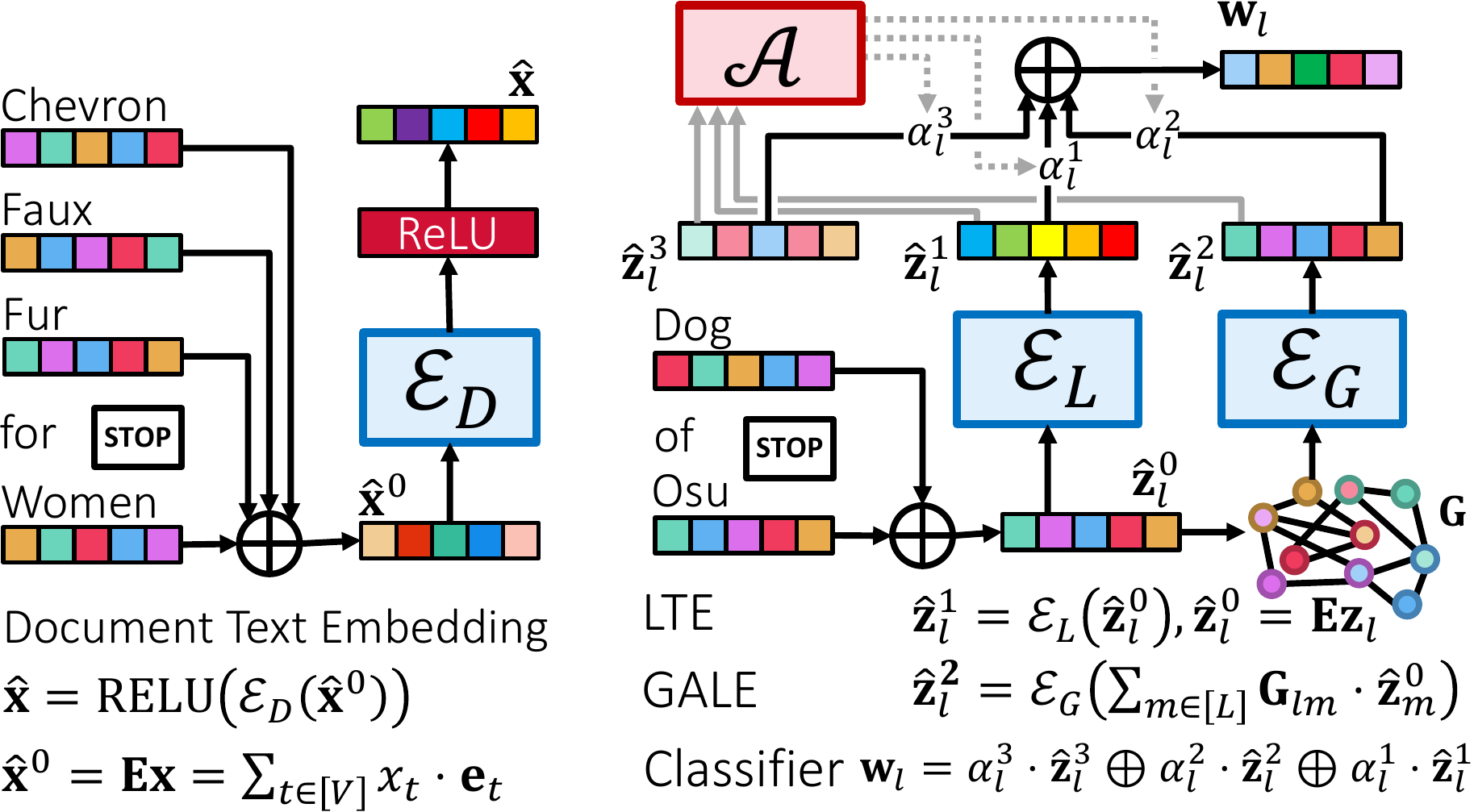}
    \caption{(Left) Document Embedding: \alg uses the light-weight embedding block $\cE$ (see Fig~\ref{fig:block}) to embed documents, ensuring rapid processing at test time (see Fig~\ref{fig:architecture}). Stop words such as \emph{for, of}) are discarded. (Right) Label classifiers: \alg incorporates multiple forms of label metadata including label text (\lt) and label correlation graphs (\ltaug), fusing them with a per-label refinement vector $\hat\vz^3_l$ using the attention block (see Fig~\ref{fig:block}) to create a \ova classifier $\vw_l$ for each label $l \in [L]$. Connections to and from the attention block are shown in light gray to avoid clutter. A separate instance of the embedding block is used to obtain document embeddings ($\cE_D$), LTE ($\cE_L$) and GALE ($\cE_G$) embeddings.}
    \label{fig:embedding}
\end{figure}

\noindent\textbf{Summary.} \alg consists of four components \textbf{1)} a text embedding architecture adapted to short-text applications, \textbf{2)} \ova classifiers, one per label that incorporate label text as well as label correlations, \textbf{3)} a shortlister that offers high-recall label shortlists for data points, allowing \alg to offer sub-millisecond prediction times even with millions of labels, and \textbf{4)} a label correlation graph that is used to train both the \ova classifiers as well as the shortlister. This section details these components as well as a technique to infer label correlation graphs from training data itself.\\

\noindent\textbf{Notation.} Let $L$ denote the number of labels and $V$ the dictionary size. All $N$ training points are presented as $(\vx_i,\vy_i)$. $\vx_i \in \bR^V$ is a bag-of-tokens representation for the $i\nth$ document i.e. $x_{it}$ is the TF-IDF weight of token $t \in [V]$ in the $i\nth$ document. $\vy_i \in \bc{-1,+1}^L$ is the ground truth label vector with $y_{il} = +1$ if label $l \in [L]$ is relevant to the $i\nth$ document and $y_{il} = -1$ otherwise. For each label $l \in [L]$, its label text is similarly represented as $\vz_l \in \bR^V$.

\subsection{Document Embedding Architecture} \alg learns $D$-dimensional embeddings for each vocabulary token $\vE = \bs{\ve_1,\ldots,\ve_V} \in \bR^{D \times V}$ and uses a light-weight embedding block (see Fig~\ref{fig:block}) implementing a residual layer. The embedding block $\cE$ contains two trainable parameters, a weight matrix $\vR$ and a scalar weight $\lambda$ (see Fig~\ref{fig:block}). Given a document $\vx \in \bR^V$ as a sparse bag-of-words vector, \alg performs a rapid embedding (see Fig~\ref{fig:embedding}) by first using the token embeddings to obtain an initial representation $\hat\vx^0 = \vE\vx \in \bR^D$, and then passing this through an instantiation $\cE_D$ of the text embedding block, and a ReLU non-linearity, to obtain the final representation $\hat\vx$. All documents (train/test) share the same embedding block $\cE_D$. Similar architectures have been shown to be well-suited to short-text applications \cite{Dahiya21,mittal20}.

\subsection{Label Correlation Graph} 
\begin{figure}
    \centering
    \includegraphics[width=0.8\columnwidth]{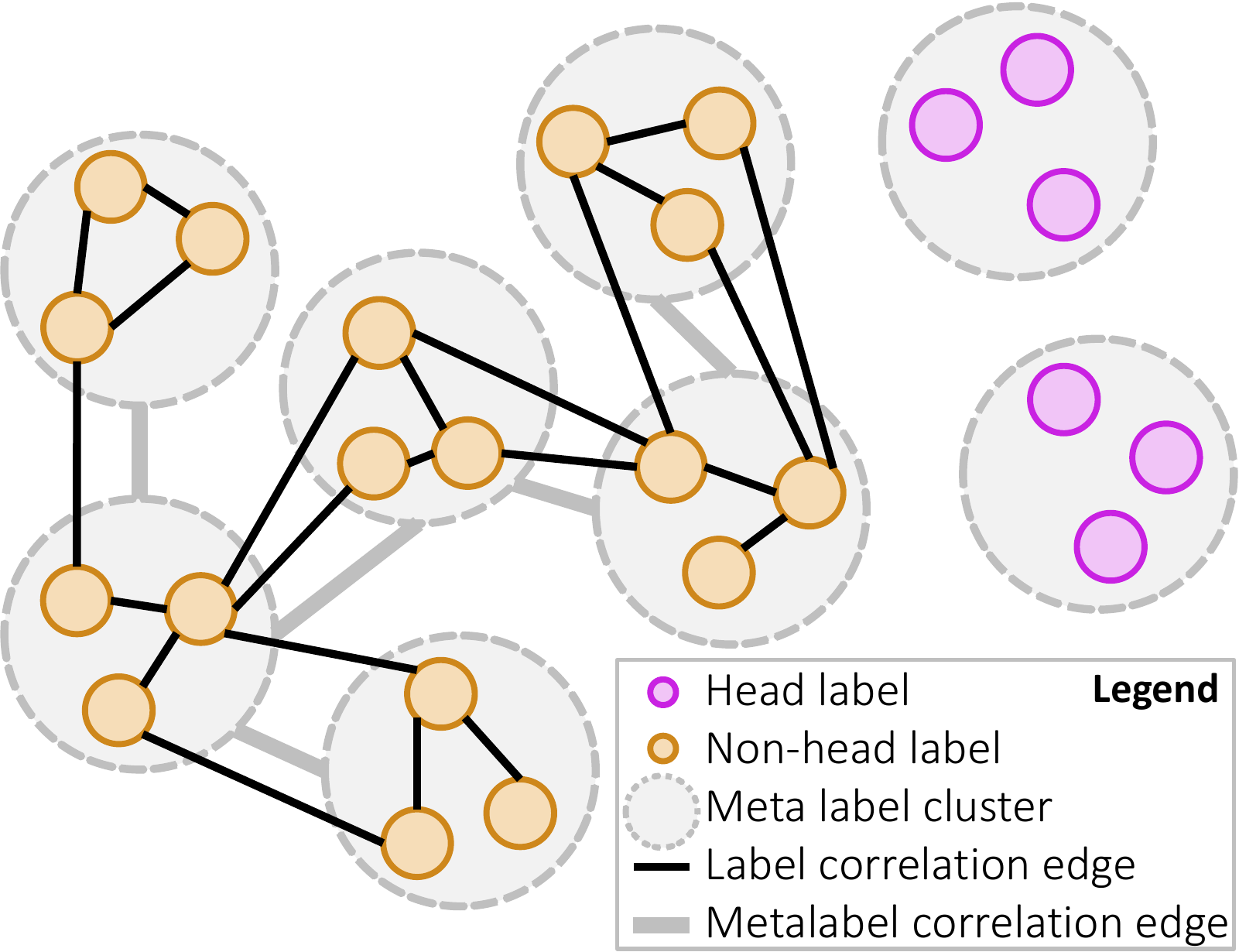}
    \caption{To avoid head labels from distorting label correlation patterns, labels with $> 500$ training points are forcibly disconnected from the label correlation graph and also clustered into \emph{head} meta-labels separately. Consequently, the \ltaug and \metaprop steps have a trivial action on head labels.}
    \label{fig:head-label}
\end{figure}

XC applications often fail to provide label correlation graphs directly as an input. Moreover, since these applications also face extreme label sparsity, using label co-occurrence alone yields fractured correlations as discussed in Sec~\ref{sec:related}. For example, label correlations gleaned from products purchased together in the same session, or else queries on which advertisers bid together, may be very sparse. To remedy this, \alg infers a label correlation graph using the ground-truth label vectors i.e. $\vy_i, i \in [N]$ themselves. This ensures that \alg is able to operate even in situations where the application is unable to provide a correlation graph itself. \alg adopts a scalable strategy based on random walks with restarts (see Algorithm \ref{algo:random}) to obtain a label correlation graph $\vG^c \in \bR^{L\times L}$ that augments the often meager label co-occurrence links (see Fig~\ref{fig:graph}) present in the ground truth. Non-rare labels (the so-called \emph{head} and \emph{torso} labels) pose a challenge to this step since they are often correlated with several labels and can overwhelm the rare labels. \alg takes two precautions to avoid this:
\begin{enumerate}[leftmargin=*]
    \item \textbf{Partition}: Head labels (those with $> 500$ training points) are disconnected from the graph by setting $\vG^c_{hh} = 1$ and $\vG^c_{ht} = 0 = \vG^c_{th}$ for all all head labels $h \in [L]$ and $t \neq h$ (see Fig~\ref{fig:head-label}).
    \item \textbf{Normalization}: $\vG^c$ is normalized to favor edges to/from rare labels as $\vG = \vA^{-1/2} \cdot \vG^c \cdot \vB^{-1/2}$, where $\vA, \vB \in \bR^{L \times L}$ are diagonal matrices with the row and column-sums of $\vG^c$  respectively.
\end{enumerate}
Algorithm \ref{algo:random} is used with a restart probability of 80\% and a random walk length of 400. Thus, it is overwhelmingly likely that several dozens of restarts would occur for each label. A high restart probability does not let the random walk wander too far thus preventing tenuous correlations among labels from getting captured.

\input{Algorithms/random_walk}

\input{figures/fig_tex/graph}

\subsection{Label Representation and Classifiers}
As examples in Tab~\ref{tab:examples} discussed in Sec~\ref{sec:results} show, label text alone may not sufficiently inform classifiers for rare labels. \alg remedies this by learning high-capacity \ova classifiers $\vW = \bs{\vw_1,\ldots,\vw_L} \in \bR^{D \times L}$ with 3 distinct components described below.\\

\noindent\textbf{Label Text Embedding (\lt).} The first component incorporates label text metadata. A separate instance $\cE_L$ of the embedding block is used to embed label text. Given a bag-of-words representation $\vz_l \in \bR^V$ of a label, the \lt representation is obtained as $\hat\vz^1_l = \cE_L(\hat\vz^0_l)$ where as before, we have the ``initial'' representation $\hat\vz^0_l = \vE\vz_l$. The embedding block $\cE_L$ is shared by all labels. We note that the DECAF method \cite{mittal20} also uses a similar architecture to embed label text.

\noindent\textbf{Graph Augmented Label Embedding (\ltaug).} \alg augments the \lt representation using the label correlation graph $\vG$ constructed earlier and a graph convolution network (GCN)~\cite{Kipf17}. This presents a departure from previous XC approaches. A typical graph convolution operation consists of two steps which \alg effectively implements at extreme scales as shown below
\begin{enumerate}[leftmargin = *]
    \item \textbf{Convolution}: initial label representations are convolved in a scalable manner using $\vG$ as $\tilde\vz^2_l = \sum_{m \in [L]} G_{lm} \cdot \hat\vz^0_m$. Note that due to random restarts used by Algorithm~\ref{algo:random}, we have $G_{ll} > 0$ for all $l \in [L]$ and thus $\tilde\vz^2_l$ contains a component from $\hat\vz^0_l$ itself.
    \item \textbf{Transformation}: Whereas traditional GCNs often use a simple non-linearity as the transformation, \alg instead uses a separate instance $\cE_G$ of the embedding block to obtain the \ltaug representation of the label as $\hat\vz^2_l = \cE_G(\tilde\vz^2_l)$.
\end{enumerate}
\alg uses a single convolution and transformation operation which allowed it to scale to applications with millions of nodes.
Recent works such as LightGCN~\cite{He2020} propose to accelerate GCNs by removing all non-linearities. Despite being scalable, using LightGCN itself was found to offer imprecise results in experiments. This may be because \alg also uses \lt representations. \alg can be seen as improving upon existing GCN architectures such as LightGCN by performing higher order label-text augmentations for $\kappa = 0,\ldots, k$ as $\hat\vz^{\kappa+1}_l = \cE^\kappa_G(\sum_{m \in [L]} G^\kappa_{lm}\cdot\hat\vz^0_m)$ where $\vG^\kappa = \prod_{j=1}^\kappa\vG$ encodes the $\kappa-$hop neighborhood, and $\cE^\kappa_G$ is a separate embedding block for each order $\kappa$. Thus, \alg's architecture allows parallelizing high-order convolutions. Whereas \alg could be used with larger orders $k > 1$, using $k = 1$ was found to already outperform all competing methods, as well be scalable.\\

\noindent\textbf{Refinement Vector and Final Classifier.} \alg combines the \lt and \ltaug representations for a label with a high-capacity per-label refinement vector $\hat\vz^3_l \in \bR^D$  (for a general value of $k$, $\hat\vz^{k+2}_l$ is used) to obtain a \ova classifier $\vw_l$ (see Fig~\ref{fig:embedding}). To combine $\hat\vz^1_l,\hat\vz^2_l,\hat\vz^3_l$, \alg uses a parameterized attention block $\cA$ (see Fig~\ref{fig:block}) to learn label-specific attention weights for the three components. This is distinct from previous works such as DECAF which use weights that are shared across labels. Fig~\ref{fig:attention} shows that \alg benefits from this flexibility, with the refinement vector $\hat\vz^3_l$ being more dominant for popular labels that have lots of training data whereas the label metadata based vectors $\hat\vz^2_l,\hat\vz^1_l$ being more important for rare labels with less training data. The attention block is explained below (see also Fig~\ref{fig:block}). Recall that \alg uses $k = 1$.
\begin{align*}
    t(\vx)&=\sigma(\vT\cdot\text{ReLU}(\vx))\\
        \vq_l &= \bs{t(\hat\vz^1_l);\ldots; t(\hat\vz^{k+2}_l)}\\
       \bs{\alpha^1_l,\ldots,\alpha^{k+2}_l} &= \frac{\exp(\vA\cdot\vq_l)}{\norm{\exp(\vA\cdot\vq_l)}_1}\\
        \vw_l &= \alpha^1_l\cdot\hat\vz^1_l + \ldots + \alpha^{k+2}_l\cdot\hat\vz^{k+2}_l
\end{align*}
The attention block is parameterized by the matrices $\vT \in \bR^{D\times D}$ and $\vA \in \bR^{(k+2) \times (k+2)D}$.
$\vq_l \in \bR^{(k+2)D}$ concatenates the transformed components before applying the attention layer $\vA$. The above attention mechanism can be seen as a scalable paramaterized option (requiring only $\bigO{D^2 + k^2D}$ additional parameters) instead of a more expensive label-specific attention scheme which would have required learning $L \times (k+2)$ parameters.

\subsection{Meta-labels and the Shortlister}
Despite being accurate, if used naively, \ova classifiers require $\Om{LD}$ time at prediction and $\Om{NLD}$ time to train. This is infeasible with millions of labels. As discussed in Sec~\ref{sec:related}, sub-linear structures are a common remedy in XC methods to perform label shortlisting during prediction \cite{Khandagale19,Prabhu18b,Yen17,Chang20,Jain19,Bhatia15,Dahiya21}. These shortlisting techniques take a data point and return a shortlist of $\bigO{\log L}$ labels that is expected to contain most of the positive labels for that data point. However, such shortlists also help during training since the negative labels that get shortlisted for a data point are arguably the most challenging and likely to get confused as being positive for that data point. Thus, \ova classifiers are trained only on positive and shortlisted negative labels, bringing training time down to $\bigO{ND\log L}$. Similar to previous works \cite{Prabhu18,mittal20}, \alg uses a clustering-based shortlister $\cS = \bc{\cC, \vH}$ where $\cC = \bc{C_1,\ldots,C_{{K}}}$ is a balanced partitioning of the $L$ labels into $K$ clusters. We refer to each cluster as a \emph{meta label}. $\vH = [\vh_1,\ldots,\vh_K] \in \bR^{D \times K}$ is a set of \ova classifiers, learnt one per meta label.\\

\noindent\textbf{Graph Assisted Multi-label Expansion (\metaprop).} \alg incorporates graph correlations to further to improve its shortlister. Let $\vM \in \bc{0,1}^{L \times K}$ denote the cluster assignment matrix i.e. $M_{lm} = 1$ if label $l$ is in cluster $m$. We normalize $\vM$ so that each column sums to unity. Given a data point $\vx$ and a \emph{beam-size} $B$, its embedding $\hat\vx$ (see Fig~\ref{fig:embedding}) is used to shortlist labels as follows
\begin{enumerate}[leftmargin = *]
    \item Find the top $B$ clusters, say $P = \bc{\tilde m_1,\ldots,\tilde m_B} \subset [K]$ according to the meta-label scores $\ip{\vh_{m_1}}{\hat\vx} \geq \ip{\vh_{m_2}}{\hat\vx} \geq \ldots$. Let $\tilde\vp \in \bR^K$ be a vector containing scores for the top $B$ clusters passed through a sigmoid i.e. $\tilde p_m = \sigma(\ip{\vh_m}{\hat\vx})$ if $m \in P$ else $\tilde p_m = 0$.
    \item Use the induced cluster-cluster correlation matrix $\vG_M = \vM^\top\vG\vM$ to calculate the ``\metaprop-ified'' scores $\vp = \vG_M\cdot\tilde\vp$.
    \item Find the top $B$ clusters according to $\vp$, say $m_1,\ldots,m_B$, retain their scores and set scores of other clusters in $\vp$ to $0$. Return the shortlist $\cS(\hat\vx) = \bc{m_1,\ldots,m_B}$ and the modified score vector $\vp$.
\end{enumerate}
Note that this cluster re-ranking step uses an induced cluster correlation graph and can bring in clusters with rare labels missed by the \ova models $\vh_m$. This is distinct from previous works which do not use label correlations for re-ranking. Since the clusters are balanced, the shortlisted clusters always contain a total of $|\cS(\hat\vx)| = LB/K$ labels. \alg uses $K = 2^{17}$ clusters and a beam size of $B \approx 30-50$ (see Sec~\ref{sec:results} for a discussion on hyperparameters).

\begin{figure}
    \centering
    \includegraphics[width=\columnwidth]{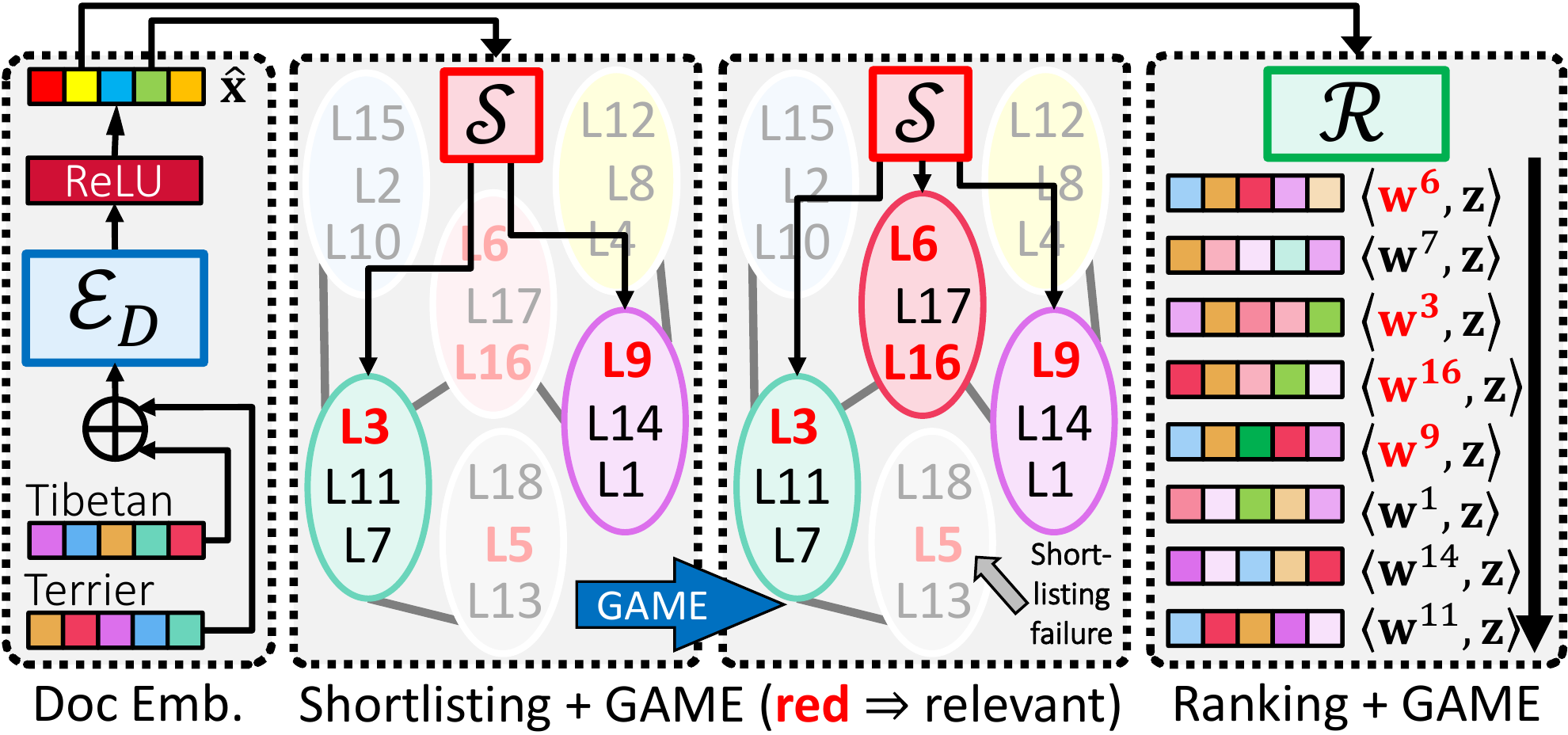}
    \caption{Prediction Pipeline: \alg uses a low-cost prediction pipeline that can make millisecond-level predictions with millions of labels. Given a document/query $\vx$, its text embedding $\hat\vx$ (see Fig~\ref{fig:embedding}) is used by the shortlister $\cS$ to obtain the $\bigO{\log L}$ most probable labels while maintaining high recall using label correlation graphs via \metaprop. Label classifiers (see Fig~\ref{fig:embedding}) for only the shortlisted labels are then used by the ranker $\cR$ to produce the final ranking of labels.}
    \label{fig:architecture}
\end{figure}

\subsection{Prediction with \metaprop}
The prediction pipeline for \alg is depicted in Fig~\ref{fig:architecture} and involves 3 steps that repeatedly utilize label correlations via \metaprop.
\begin{enumerate}
    \item Given a document $\vx$, use the shortlister $\cS$ to get a set of $B$ meta labels $\cS(\hat\vx)$ and their corresponding scores $\vp \in \bR^K$.
    \item For shortlisted labels, apply the \ova classifiers $\vw_l$ to calculate the ranking score vector $\tilde\vr \in \bR^L$ with $\tilde r_l = \sigma(\ip{\vw_l}{\hat\vx})$ if $l \in C_m$ for some $m \in \cS(\hat\vx)$ else $\tilde r_l = 0$.
    \item \metaprop-ify the \ova scores to get $\vr = \vG\cdot\tilde\vr$ (note that $\vG$ is used this time and not $\vG_M$). Make final predictions using the joint scores $s_l := r_l \cdot p_m$ if $l \in C_m, m \in \cS(\hat\vx)$ else $s_l = 0$.
\end{enumerate}

\subsection{Efficient Training: the DeepXML Pipeline}
\noindent\textbf{Summary.} \alg adopts the scalable DeepXML pipeline \cite{Dahiya21} that splits training into 4 modules. In summary, Module I jointly learns the token embeddings $\vE$, the embedding block $\cE_D$ and the shortlister $\cS$. $\vE$ remains frozen hereon. Module II refines $\cP$ and uses it to retrieve label shortlists for all data points. After performing initialization in Module III, Module IV uses the shortlists generated in Module II to jointly fine-tune $\cE_D$ and learn the \ova classifiers $\vW$, implicitly learning the embedding blocks $\cE_L, \cE_G$ and the attention block $\cA$ in the process.\\

\noindent\textbf{Module I.} Token embeddings $\vE \in \bR^{D \times V}$ are randomly initialized using~\cite{he2015delving}, the residual block within $\cE_D$ is initialized to identity. After creating label clusters (see below), each cluster is treated as a \emph{meta-label} yielding a \emph{meta}-XC problem on the same training points, but with $K$ meta-labels instead of the original $L$ labels. Meta-label text is created for each $m \in [K]$ as $\vu_m = \sum_{l \in C_m}\vz_l$. Meta labels are also endowed with the meta-label correlation graph $\vG_M = \vM^\top\vG\vM$ where $\vM \in \bc{0,1}^{L \times K}$ is the cluster assignment matrix. \Ova meta-classifiers $\vH = [\vh_1,\ldots,\vh_K] \in \bR^{D \times K}$ are now learnt to solve this meta XC problem. These classifiers have the same form as those for the original problem with 3 components, \lt, \ltaug, (with corresponding blocks $\tilde\cE_L,\tilde\cE_G$) and refinement vector, with an attention block $\tilde\cA$ supervising their combination (parameters within $\tilde\cA$ are initialized randomly). However, in Module-I, refinement vectors are turned off for $\vh_l$ to force good token embeddings $\vE$ to be learnt without support from refinement vectors. Module I solves the meta XC problem while training $\cE_D, \tilde\cE_L,\tilde\cE_G, \tilde\cA, \vE$, implicitly learning $\vH$.\\

\noindent\textbf{Meta-label Creation with Graph Augmented Label Centroids.} Existing approaches such as Parabel or DECAF cluster labels by creating a label centroid for each label $l$ by aggregating features for training documents associated with that label as $\vc_l = \sum_{i:y_{il} = +1}\vx_i$. However, \alg recognizes that missing labels often lead to an incomplete ground truth, and thus, poor label centroids for rare labels in XC settings~\cite{Jain17, Babbar19}. Fig~\ref{fig:clusters} confirms this suspicion. \alg addresses this by augmenting the label centroids using the label co-occurrence graph $\vG$ to redefine the centroids as $\hat\vc_l = \sum_{j=1}^{L} G_{lj}\cdot\vc_l$. Balanced hierarchical binary clustering \cite{Prabhu18b} is now done on these label centroids for 17 levels to generate $K = 2^{17}$ label clusters. Note that since token embeddings have not been learnt yet, the raw TF-IDF documents vectors $\vx_i \in \bR^V$ are used instead of $\hat\vx_i \in \bR^D$.\\

\noindent\textbf{Module II.} The shortlister is fine-tuned in this module. Label centroids are recomputed as $\hat\vc_l = \sum_{j=1}^{L} G_{lj}\cdot\vc_l$ where this time, $\vc_l = \sum_{i:\vy^i_l = +1}\vE\vx_i$ using $\vE$ learnt in Module I. The meta XC problem is recreated and solved again. However, this time, \alg allows the meta-classifiers $\vh_m$ to also include refinement vectors to better solve the meta-XC problem. In the process of re-learning $\vH$, the model parameters $\cE_D, \tilde\cE_L, \tilde\cE_G, \tilde\cA$ are fine-tuned ($\vE$ is frozen after Module I). The shortlister $\cS$ thus obtained is thereafter used to retrieve shortlists $\cS(\hat\vx_i)$ for each data point $i \in [N]$. However, distinct from previous work such as Slice~\cite{Jain19}, X-Transformer~\cite{Chang20}, DECAF~\cite{mittal20}, \alg uses the \metaprop strategy to obtain negative samples that take label correlations into account.\\

\noindent\textbf{Module III.} Residual blocks within $\cE_D, \cE_L, \cE_G$ are initialized to identity, parameters within $\cA$ are initialized randomly, and the shortlister $\cS$ and token embeddings $\vE$ are kept frozen. Refinement vectors for all $L$ labels are initialized to $\hat\vz^3_l = \sum_{m \in [L]} G_{lm} \cdot \vE\vz_m$. We find this initialization to be both crucial (see Sec~\ref{sec:results}) as well as distinct from previous works such as DECAF which initialized its counterpart of refinement vectors using simply $\vE\vz_l$. Such correlation-agnostic initialization was found to offer worse results than \alg's graph augmented initialization.\\

\noindent\textbf{Module IV.} In this module, the embedding blocks $\cE_D, \cE_L, \cE_G$ are learnt jointly with the per-label refinement vectors $\hat\vz^3_l$ and attention block $\cA$, thus learning the \ova classifiers $\vw_l$ in the process, However, training is done in $\bigO{ND\log L}$ time by restricting training to positives and shortlisted negatives for each data point.\\

\noindent\textbf{Loss Function and Regularization.} \alg uses the binary cross entropy loss function for training in Modules I, II and IV using the Adam~\cite{Kingma14} optimizer. The residual weights $\vR$ in the various embedding blocks $\cE_D, \cE_L, \cE_G, \tilde\cE_L, \tilde\cE_G$ as well as the weights $\vT$ in the attention block were all subjected to spectral regularization \cite{Miyato18}. All ReLU layers in the architecture also included a dropout layer with 20\% rate.\\

\noindent\textbf{Key Contributions in Training.} \alg markedly departs from existing techniques by incorporating label correlation information in a scalable manner at every step of the learning process. Right from Module I, label correlations are incorporated while creating the label centroids leading to higher quality clusters (see Fig~\ref{fig:clusters} and Table~\ref{tab:lmi}). The architecture itself incorporates label correlation information using the \ltaug representations. It is crucial to initialize the refinement vectors $\hat\vz^3_l$ properly for which \alg uses a graph-augmented initialization. \alg continues infusing label correlation during negative sampling using the \metaprop step. Finally, \metaprop is used multiple times in the prediction pipeline as well. As the discussion in Sec~\ref{sec:results} will indicate, these augmentations are crucial to the performance boosts offered by \alg.

%% file: Algorithms/random_walk.tex
\begin{algorithm}[t]
\caption{\textbf{Label Correlation Graph Genration}. $N, L$ denote the number of training points and labels. $\omega$ and $p$ denote the walk length and restart probability. $\vY = [\vy_1,\ldots,\vy_N] \in \bc{-1,+1}^{L \times N}$ is a matrix giving the relevant training labels for each document. For any set $S$, $\unif(S)$ returns a uniformly random sample from $S$. The subroutine \textsc{WalkFrom} performs a walk starting at the label $l$.}
\begin{algorithmic}[1]
\Procedure{WalkFrom}{$l, \vY, \omega, p$}
\State $\vv \gets \vzero \in \bR^L$ \Comment{Initialize visit counts}
\State $\lambda \gets l$ \Comment{Start the walk from the label $l$}
\For{$t = 1; t \leq \omega; t$++}
    \State $\phi \gets \unif([0, 1])$\; \Comment{Random number in range $[0,1]$}
    \If{$\phi \leq p$}
        \State $\lambda = l$\; \Comment{Restart if required}
    \EndIf
    \State $\delta \gets \unif(\bc{i: Y_{\lambda i} = +1})$\; \Comment{Sample a relevant doc}
    \State $\lambda \gets \unif(\bc{j: Y_{j\delta} = +1})$\; \Comment{Sample a relevant label}
    \State $\vv[\lambda]$++\; \Comment{Update the visit counts}
\EndFor
\State \textbf{return} $\vv$
\EndProcedure

\Procedure{RandomWalk}{$L, \vY, \omega, p$}
\State $\vG^c \gets \vzero\vzero^\top \in \bR^{L \times L}$\; \Comment{Initialize visit counts}
\For{$l = 1; l \leq L; l$++}
    \State $\vG_l^c \gets \text{\textsc{WalkFrom}}(l, \vY, \omega, p)$\; \Comment{Update row $l$ of $\vG^c$}
\EndFor
\State \textbf{return} $\vG^c$\;
\EndProcedure
\end{algorithmic}
\label{algo:random}
\end{algorithm}

%% file: figures/fig_tex/graph.tex
\begin{figure}[t]
    \centering
    \includegraphics[width=\linewidth]{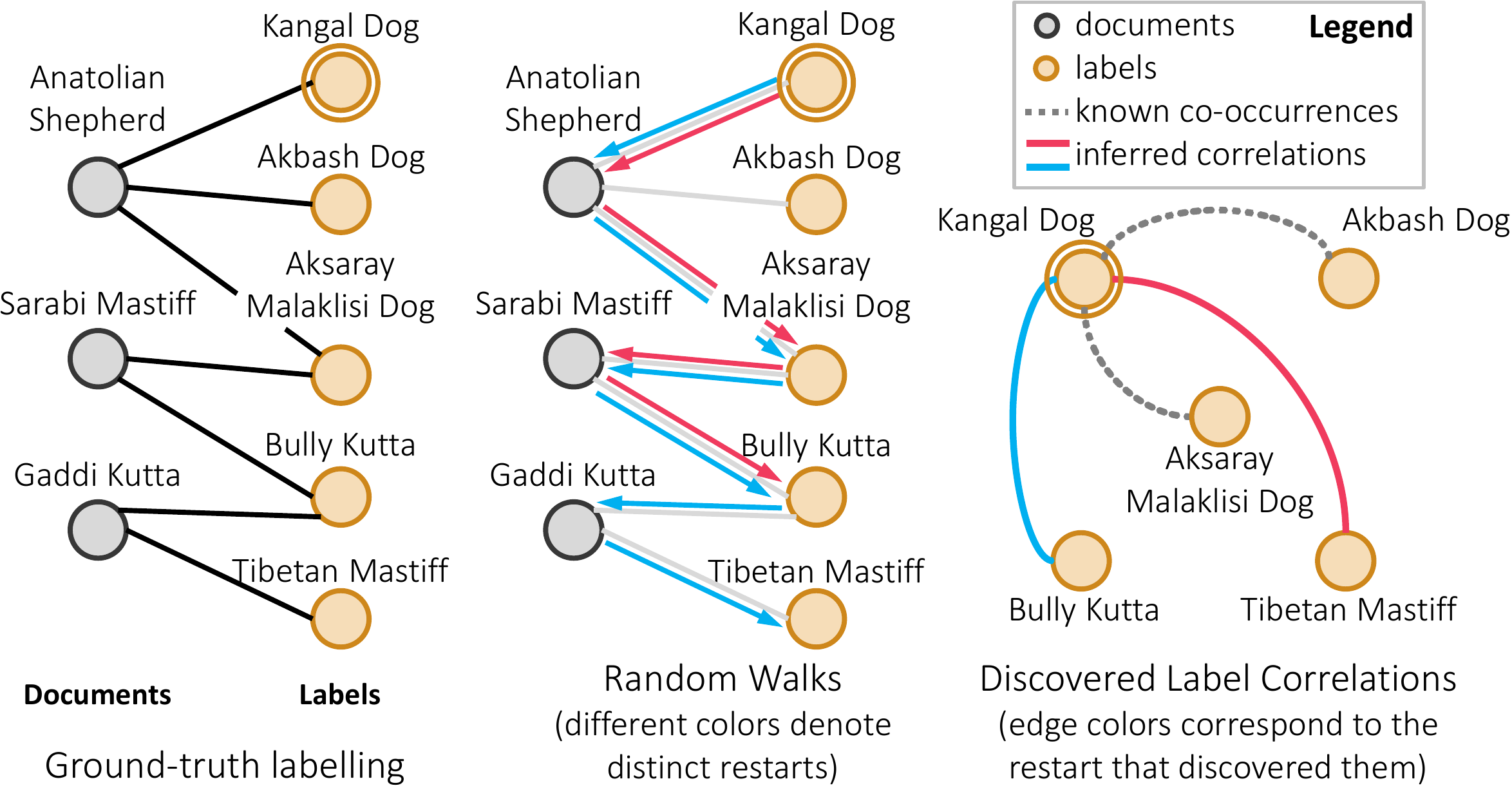}
    \caption{An execution of Algorithm~\ref{algo:random} on a subset of the LF-WikiSeeAlsoTitles-320K dataset starting from the label ``Kangal Dog''. (Left) \alg uses document-label associations taken from ground-truth label vectors (black color) to infer indirect correlations among labels. (Middle) Random walks (distinct restarts colored differently) infer diverse correlations. (Right) These inferred correlations (marked with the color of the restart that discovered them) augment the meager label co-occurrences present in the ground truth (marked with dotted lines).}
    \label{fig:graph}
\end{figure}

%% file: sections/experiments.tex
\section{Experiments} \label{sec:results}
\input{tables/datastats}
\input{tables/results_main_filtered_titles}
\input{tables/P2P}
\input{tables/knn_graph}

\noindent\textbf{Datasets and Features.} \alg was evaluated on 4 publicly available\footnote{Extreme Classification Repository \cite{XMLRepo}} benchmark datasets, LF-AmazonTitles-131K, LF-WikiSeeAlso-Titles-320K, LF-WikiTitles-500K and LF-AmazonTitles-1.3M. These datasets were derived from existing datasets e.g. Amazon-670K, by taking those labels for which label text was available and performing other sanitization steps such as reciprocal pair removal (see \cite{mittal20} for details). \alg was also evaluated on proprietary datasets P2P-2M and P2P-10M, both mined from click logs of the Bing search engine, where a pair of products were considered similar if the Jaccard index of the set of queries which led to a click on them was found to be more than a certain threshold. \alg used the word piece tokenizer~\cite{schuster12} to create a shared vocabulary for documents and labels. Please refer to Tab~\ref{tab:stats} for dataset statistics. \\

\noindent\textbf{Baseline algorithms.} \alg's performance was compared to state-of-the-art deep extreme classifiers which jointly learn document and label representations such as DECAF~\cite{mittal20}, AttentionXML \cite{You18}, Astec~\cite{Dahiya21}, X-Transformer~\cite{Chang20}, and MACH~\cite{Medini2019}. DECAF and X-Transformer are the only methods that also use label text and are therefore the most relevant for comparison with \alg. For the sake of completeness, \alg was also compared to classifiers which use fixed document representations like DiSMEC~\cite{Babbar17}, Parabel~\cite{Prabhu18b}, Bonsai~\cite{Khandagale19}, and Slice~\cite{Jain19}. All these fixed-representation methods use BoW features, except Slice which used pre-trained FastText \cite{Bojanowski17} features. The GLaS~\citep{Guo2019} method could not be included in our analysis as its code was not publicly available. \\

\noindent\textbf{Evaluation.} Methods were compared using standard XC metrics, namely Precision~(P@$k$) and Propensity-scored precision~(PSP@$k$)~\cite{Jain17}. Recall~(R@$K$) was also included since XC methods are typically used in the shortlisting pipeline of recommendation systems. Thus, having high recall is equally important as having high precision. For evaluation, guidelines provided on the XML repository~\citep{XMLRepo} were followed. To be consistent, all models were run on a 6-core Intel Skylake 2.4 GHz machine with one Nvidia V100 GPU.\\

\noindent\textbf{Hyper-parameters.} A beam size of $B = 30$ was used for the dataset LF-AmazonTitles-131K and $B = 50$ for all other datasets. Embedding dimension $D$ was set to 300 for datasets with $< 400$K labels and 512 otherwise. The number of meta-labels $K = |\cC|$ was fixed to $|\cC|=2^{17}$ for all other datasets except for LF-AmazonTitles-131K where $|\cC|=2^{15}$ was chosen since the dataset itself has around $2^{17}$ labels. The default PyTorch implementation of the Adam optimizer was used. Dropout with probability 0.2 was used for all datasets. Learning rate was decayed by a decay factor of 0.5 after an interval of $0.5\times$ epoch length. Batch size was taken to be 255 for all datasets. \alg Module-I used 20 epochs with an initial learning rate of 0.01. In Modules-II and IV, 10 epochs were used for all datasets with an initial learning rate of 0.008. While constructing the label correlation graph using random walks (see Algorithm~\ref{algo:random}), a walk length of $400$ and restart probability of $80\%$ were used for all datasets.\\

\input{figures/fig_tex/results_contrib}

\noindent\textbf{Results on benchmark datasets.} Tab~\ref{tab:baelines_eval_p2p} demonstrates that \alg can be significantly more accurate than existing XC methods. In particular, \alg could be upto 3\% and 10\% more accurate as compared to DECAF and X-Transformer respectively in terms of P@$1$. It should be noted that DECAF and X-Transformer are the only XC methods which use label meta-data. Furthermore, \alg could outperform Astec which is specifically designed for rare labels, by up to 7\% in terms of PSP@$1$, indicating that \alg offers state-of-the-art accuracies without compromising on rare labels. To further understand  the gains of \alg, the labels were divided into five bins such that each bin contained an equal number of positive training points (Fig~\ref{fig:sup:contrib}). This ensured that each bin had an equal opportunity to contribute to the overall accuracy. Fig~\ref{fig:sup:contrib} indicates that the major gains of \alg come from predicting rare labels correctly. \alg could outperform all other deep-learning-based XC methods, as well as fixed-feature-based XC methods by a significant margin of at least 7\%. Moreover, \alg's recall could be up to 2\% higher as compared to all other XC methods.\\

\noindent\textbf{Results on Propreitary Bing datasets.} \alg's performance was also compared on the proprietary datasets LF-P2PTitles-2M and LF-P2PTitles-10M. Note that \alg was only compared with those XC methods that were able to scale to 10 million labels on a single GPU within a timeout of one week. \alg could be up to 14\%, 15\%, and 15\% more accurate as compared to state-of-the-art methods in terms of P@$1$, PSP@$1$ and R@$10$ respectively. Please refer to Tab~\ref{tab:p2p} for more details.\\

\input{tables/ablation}

\noindent\textbf{Ablation experiments.} 
To scale accurately to millions of labels, \alg makes several meticulous design choices. To validate their importance, Tab~\ref{tab:ablation} compares different variants of \alg:
\begin{enumerate}[leftmargin=*]
    \item \textbf{Label-graph augmentations:} The label correlation graph $\vG$ is used by \alg in several steps, such as in creating meaningful meta labels, negative sampling, shortlisting of meta-labels, \etc. To evaluate the importance of these augmentations, in \alg-NoGraph, the graph convolution component $\vG$ was removed from all training steps such as \metaprop, except while generating the label classifiers (\ltaug). \alg-NoGraph was found to be up to 3\% less accurate than \alg. Furthermore, in a variant \alg-PPR, inspired by state-of-the-art graph algorithms~\cite{ying2018, klicpera18}, negative sampling was performed via Personalized Page Rank. \alg could be up to 25\% more accurate than \alg-PPR. This could be attributed to the highly sparse label correlation graph and justifies the importance of \alg's label correlation graph as well as it's careful negative sampling.

    \item \textbf{Label components:} Much work has been done to augment (document) text representations using graph convolution networks (GCNs)~\cite{Tang20, Kipf17, zhang20, hamilton17, He2020, velickovic18, pal20}. These methods could also be adapted to convolve label text for \alg's \ltaug component. \alg's \ltaug component was replaced by a LightGCN~\cite{He2020} and GCN~\cite{Kipf17} (with the refinement vectors in place). Results indicate that \alg could be upto 2\% and 10\% more accurate as compared to LightGCN and GCN. In another variant of \alg, the refinement vectors $\hat\vz^3_l$ were removed (\alg-NoRefine). Results indicate that \alg could be up to 10\% more accurate as compared to \alg-NoRefine which indicates that the per-label (extreme) refinement vectors are essential for accuracy.
    
    \item \textbf{Graph construction:} To evaluate the efficacy of \alg's graph construction, we compare it to \alg-Cooc where we consider the label co-occurrence graph generated by $\vY^\top\vY$ instead of the random-walk based graph $\vG$ used by \alg. \alg-Cooc could be up to 2\% less accurate in terms of PSP@1 than \alg. This shows that \alg's random-walk graph indeed captures long-term dependencies thereby resulting in improved performance on rare labels. Normalizing a directed graph, such as the graph $\vG^c$ obtained by Algorithm~\ref{algo:random} is a non-trivial problem. In \alg-PN, we apply the popular Perron normalization~\cite{chung05} to the random-walk graph $\vG^c$. Unfortunately, \alg-PN leads to a significant loss in propensity-scored metrics for rare labels. This validates the choice of \alg's normalization strategy.
    
    \item \textbf{Combining label representations:} The \lt and \ltaug components of \alg could potentially be combined using strategies different from the attention mechanism used by \alg. A simple average/sum of the components, (\alg-SUM) could be up to 2\% worse, which corroborates the need for the attention mechanism for combining heterogeneous components.
    
    \item \textbf{Higher-order Convolutions:} Since the \alg framework could handle higher order convolutions $k > 1$ efficiently, we validated the effect of increasing the order. Using $k = 2$ was found to hurt precision by upto 2\%. Higher orders \eg. $k = 3, 4$ \etc. were intractable at XC scales as the graphs got too dense. The drop in performance when using $k = 2$ could be due to two reasons: (a) at XC scales, exploring higher order neighborhoods would add more noise than information unless proper graph pruning is done afterward and (b) \alg's random-walk graph creation procedure already encapsulates some form of higher order proximity which negates the potential for massive benefits when using higher order convolutions.
    
    \item \textbf{Meta-labels and \metaprop}: \alg uses a massive fanout of $|\cC|=2^{17}$ meta-labels in its shortlister. Ablations with $|\cC|=2^{13}$ (\alg-8K) show that using a smaller fanout can lead to upto 8\% loss in precision. Additionally Tab~\ref{tab:knngraph} shows that incorporating \metaprop with other XC methods can also improve their accuracies (although \alg continues to lead by a wide margin). In particular incorporating \metaprop in Parabel and AttentionXML led to up to 1\% increase in accuracy. This validates the utility of \metaprop as a general XC tool.
\end{enumerate}

\input{figures/fig_tex/clusters}
\input{figures/fig_tex/figure_attention}
\input{tables/examples}
\input{tables/LMI}
\noindent\textbf{Analysis.} This section critically analyzes \alg's performance gains by scrutinizing the following components:
\begin{enumerate}[leftmargin=*]
    \item \textbf{Clustering:} As is evident from  Fig~\ref{fig:clusters}, \alg's offers significantly better clustering quality. Other methods such as DECAF use label centroids over an incomplete ground truth, resulting in clusters of seemingly unrelated labels. For \eg. the label ``\textsf{Bulldog}'' was clustered with ``\textsf{Great house at Sonning}'' by DECAF and the label \notsametext was clustered with ``\textsf{Ferdinand II of Aragon}'' which never co-occur in training. However \alg clusters much more relevant labels together, possibly since it was able to (partly) complete the ground truth using it's label correlation graph $\vG$. This was also verified quantitatively by evaluating the clustering quality using the standard Loss of Mutual Information metric (LMI)~\cite{dhillon03}. Tab~\ref{tab:lmi} shows that \alg has the least LMI compared to other methods such as DECAF and those such as MACH that use random hashes to cluster labels. 

    \item \textbf{Component Contribution:} \alg chooses to dynamically attend on multiple (and heterogeneous) label representations in its classifier components, which allows it to capture nuanced variations in the semantics of each label. To investigate the contribution of the label text and refinement classifiers, Fig.~\ref{fig:attention} plots the average product of the attention weight and norm of each component. It was observed that the label text components \lt and \ltaug are crucial for rare labels whereas the (extreme) refinement vector $\hat\vz^3_l$ is more important for data-abundant popular labels. 

    \item \textbf{Label Text Augmentation}: For the document \testdocument, \alg could make correct rare label predictions like \notsametext even when the label text exhibits no token similarity with the document text or other co-occurring labels. Other methods such as DECAF failed to understand the semantics of the label and mis-predicted the label ``\textsf{Fox Terrier}'' wrongly relying on the token ``\textsf{Terrier}''. We attribute this gain to \alg's label correlation graph as \notsametext correlated well with the labels \co in $\vG$. Several such examples exist in the datasets. In another example from the P2P-2M dataset, for the product ``\textsf{Draper's \& Damon's Women's  Chevron Fauz Fur Coat Tan L}'', \alg could deduce the intent of purchasing ``\textsf{Fur Coat}'' while other XC methods incorrectly fixated on the brand ``\textsf{Draper's \& Damon's}''. Please refer to Tab~\ref{tab:examples} for detailed examples.
\end{enumerate}

%% file: tables/datastats.tex
\begin{table*}
	\caption{Dataset Statistics. A $\ddagger$ sign denotes information that was redacted for proprietary datasets. The first four rows are public datasets and the last two rows are proprietary datasets. Dataset names with an asterisk $^\ast$ next to them correspond to product-to-category tasks whereas others are product-to-product tasks.}
	\label{tab:stats}
	\resizebox{\linewidth}{!}
	{
		\begin{tabular}{l|cccccccc}
			\toprule
			\textbf{Dataset} &
			\textbf{\begin{tabular}[c]{@{}c@{}}Train Documents \\ $N$ \end{tabular}} &
			\textbf{\begin{tabular}[c]{@{}c@{}}Labels\\ $L$ \end{tabular}}  &
			\textbf{\begin{tabular}[c]{@{}c@{}}Tokens\\ $V$ \end{tabular}} &
			\textbf{\begin{tabular}[c]{@{}c@{}}Test Instances \\ $N'$\end{tabular}} &
			\textbf{\begin{tabular}[c]{@{}c@{}}Average Labels\\ per Document \end{tabular}} &
			\textbf{\begin{tabular}[c]{@{}c@{}}Average Points\\ per label \end{tabular}} &
			\textbf{\begin{tabular}[c]{@{}c@{}}Average Tokens\\ per Document \end{tabular}} &
			\textbf{\begin{tabular}[c]{@{}c@{}}Average Tokens\\ per Label \end{tabular}} \\
			\midrule
			\multicolumn{9}{c}{Short text dataset statistics}\\ \midrule
            LF-AmazonTitles-131K & 294,805 & 131,073 & 40,000 & 134,835 & 2.29 & 5.15 & 7.46 & 7.15 \\
            LF-WikiSeeAlsoTitles-320K & 693,082 & 312,330 & 40,000 & 177,515 & 2.11 & 4.68 & 3.97 & 3.92 \\
            LF-WikiTitles-500K$^\ast$ & 1,813,391 & 501,070 & 80,000 & 783,743 & 4.74 & 17.15 & 3.72 & 4.16 \\
            LF-AmazonTitles-1.3M & 2,248,619 & 1,305,265 & 128,000 & 970,237 & 22.20 & 38.24 & 9.00 & 9.45 \\\midrule
            \midrule
            \multicolumn{9}{c}{Proprietary dataset}\\
            \midrule
            LF-P2PTitles-2M & 2,539,009 & 1,640,898 & $\ddagger$ & 1,088,146 & $\ddagger$ & $\ddagger$ & $\ddagger$ & $\ddagger$ \\
            LF-P2PTitles-10M & 6,849,451 & 9,550,772 & $\ddagger$ & 2,935,479 & $\ddagger$ & $\ddagger$ & $\ddagger$ & $\ddagger$ \\
            \bottomrule
    	\end{tabular}
	}
\end{table*}

%% file: tables/results_main_filtered_titles.tex
\begin{table}[t]
    \caption{Results on public benchmark datasets. \alg could offer 2-3.5\% higher P@1 as well as upto 5\% higher PSP@1 which focuses on rare labels. Additionally, \alg offered up to 3\% better recall than leading XC methods.}
    \label{tab:baelines_eval_p2p}
      \centering
      \resizebox{!}{0.43\textheight}{
        \begin{tabular}{@{}l|cc|cc|c|c@{}}
        \toprule
        \textbf{Method} & \textbf{PSP@1} & \textbf{PSP@5} & \textbf{P@1} & \textbf{P@5} & \textbf{R@10} & \textbf{\begin{tabular}[c]{@{}c@{}} Prediction \\ Time (ms) \end{tabular}} \\
        \midrule
\multicolumn{7}{c}{LF-AmazonTitles-131K}\\ \midrule						
\alg	 & \textbf{33.51}	 & \textbf{44.7}	 & \textbf{40.74}	 & \textbf{19.88}	 & \textbf{54.11}	 & 0.1 \\
DECAF	 & 30.85	 & 41.42	 & 38.4	 & 18.65	 & 51.2	 & 0.1 \\
Astec	 & 29.22	 & 39.49	 & 37.12	 & 18.24	 & 49.87	 & 2.34 \\
AttentionXML	 & 23.97	 & 32.57	 & 32.25	 & 15.61	 & 42.3	 & 5.19 \\
Slice	 & 23.08	 & 31.89	 & 30.43	 & 14.84	 & 41.16	 & 1.58 \\
MACH	 & 24.97	 & 34.72	 & 33.49	 & 16.45	 & 44.75	 & 0.23 \\
X-Transformer	 & 21.72	 & 27.09	 & 29.95	 & 13.07	 & 35.59	 & 15.38 \\
Siamese	 & 13.3	 & 13.36	 & 13.81	 & 5.81	 & 14.69	 & 0.2 \\
Bonsai	 & 24.75	 & 34.86	 & 34.11	 & 16.63	 & 45.17	 & 7.49 \\
Parabel	 & 23.27	 & 32.14	 & 32.6	 & 15.61	 & 41.63	 & 0.69 \\
DiSMEC	 & 25.86	 & 36.97	 & 35.14	 & 17.24	 & 46.84	 & 5.53 \\
XT	 & 22.37	 & 31.64	 & 31.41	 & 15.48	 & 42.11	 & 9.12 \\
AnneXML	 & 19.23	 & 32.26	 & 30.05	 & 16.02	 & 45.57	 & 0.11 \\
\midrule
\multicolumn{7}{c}{LF-WikiSeeAlsoTitles-320K}\\ \midrule						
\alg	 & \textbf{22.01}	 & \textbf{26.27}	 & \textbf{29.35}	 & \textbf{15.05}	 & \textbf{36.46}	 & 0.12 \\
DECAF	 & 16.73	 & 21.01	 & 25.14	 & 12.86	 & 32.51	 & 0.09 \\
Astec	 & 13.69	 & 17.5	 & 22.72	 & 11.43	 & 28.18	 & 2.67 \\
AttentionXML	 & 9.45	 & 11.73	 & 17.56	 & 8.52	 & 20.56	 & 7.08 \\
Slice	 & 11.24	 & 15.2	 & 18.55	 & 9.68	 & 24.45	 & 1.85 \\
MACH	 & 9.68	 & 12.53	 & 18.06	 & 8.99	 & 22.69	 & 0.52 \\
Siamese	 & 10.1	 & 9.59	 & 10.69	 & 4.51	 & 10.34	 & 0.17 \\
Bonsai	 & 10.69	 & 13.79	 & 19.31	 & 9.55	 & 23.61	 & 14.82 \\
Parabel	 & 9.24	 & 11.8	 & 17.68	 & 8.59	 & 20.95	 & 0.8 \\
DiSMEC	 & 10.56	 & 14.82	 & 19.12	 & 9.87	 & 24.81	 & 11.02 \\
XT	 & 8.99	 & 11.82	 & 17.04	 & 8.6	 & 21.73	 & 12.86 \\
AnneXML	 & 7.24	 & 11.75	 & 16.3	 & 8.84	 & 23.06	 & 0.13 \\
\midrule
\multicolumn{7}{c}{LF-AmazonTitles-1.3M}\\ \midrule						
\alg	 & \textbf{23.43}	 & \textbf{30.56}	 & 50.14	 & 40	 & \textbf{32.02}	 & 0.32 \\
DECAF	 & 22.07	 & 29.3	 & \textbf{50.67}	 & \textbf{40.35}	 & 31.29	 & 0.16 \\
Astec	 & 21.47	 & 27.86	 & 48.82	 & 38.44	 & 29.7	 & 2.61 \\
AttentionXML	 & 15.97	 & 22.54	 & 45.04	 & 36.25	 & 26.26	 & 29.53 \\
Slice	 & 13.96	 & 19.14	 & 34.8	 & 27.71	 & 20.21	 & 1.45 \\
MACH	 & 9.32	 & 13.26	 & 35.68	 & 28.35	 & 19.08	 & 2.09 \\
Bonsai	 & 18.48	 & 25.95	 & 47.87	 & 38.34	 & 29.66	 & 39.03 \\
Parabel	 & 16.94	 & 24.13	 & 46.79	 & 37.65	 & 28.43	 & 0.89 \\
DiSMEC	 & -	 & -	 & -	 & -	 & -	 & - \\
XT	 & 13.67	 & 19.06	 & 40.6	 & 32.01	 & 22.51	 & 5.94 \\
AnneXML	 & 15.42	 & 21.91	 & 47.79	 & 36.91	 & 26.79	 & 0.12 \\
\midrule
\multicolumn{7}{c}{LF-WikiTitles-500K}\\ \midrule						
\alg	 & \textbf{21.58}	 & 19.84	 & 44.36	 & 16.91	 & 30.59	 & 0.14 \\
DECAF	 & 19.29	 & \textbf{19.96}	 & 44.21	 & 17.36	 & \textbf{32.02}	 & 0.09 \\
Astec	 & 18.31	 & 18.56	 & \textbf{44.4}	 & \textbf{17.49}	 & 31.58	 & 2.7 \\
AttentionXML	 & 14.8	 & 13.88	 & 40.9	 & 15.05	 & 25.8	 & 9 \\
Slice	 & 13.9	 & 13.82	 & 25.48	 & 10.98	 & 22.65	 & 1.76 \\
MACH	 & 13.71	 & 12	 & 37.74	 & 13.26	 & 23.81	 & 0.8 \\
Bonsai	 & 16.58	 & 16.4	 & 40.97	 & 15.66	 & 28.04	 & 17.38 \\
Parabel	 & 15.55	 & 15.35	 & 40.41	 & 15.42	 & 27.34	 & 0.81 \\
DiSMEC	 & 15.88	 & 15.89	 & 39.42	 & 14.85	 & 26.73	 & 11.71 \\
XT	 & 14.1	 & 14.38	 & 38.13	 & 14.66	 & 26.48	 & 7.56 \\
AnneXML	 & 13.91	 & 13.75	 & 39	 & 14.55	 & 26.27	 & 0.13 \\
        \bottomrule
    \end{tabular}}
\end{table}

%% file: tables/P2P.tex
\begin{table}[ht]
    \caption{Results on proprietary product-to-product (P2P) recommendation datasets. \alg could offer significant gains -- upto 14\% higher P@1, 15\% higher PSP@1 and 7\% higher R@10 -- than competing classifiers.}
    \label{tab:p2p}
      \centering
      \resizebox{\linewidth}{!}{
       \begin{tabular}{@{}l|ccc|ccc|c@{}}
        \toprule
        \textbf{Method} & \textbf{PSP@1} & \textbf{PSP@3} & \textbf{PSP@5}& \textbf{P@1} & \textbf{P@3} & \textbf{P@5} & \textbf{R@10} \\
        \midrule
        \multicolumn{8}{c}{LF-P2PTitles-2M}\\ \midrule							
    \alg	 & \textbf{41.97}	 & \textbf{44.92}	 & \textbf{49.46}	 & \textbf{43.79}	 & \textbf{39.25}	 & \textbf{33.15}	 & \textbf{54.44} \\
    DECAF	 & 36.65	 & 40.14	 & 45.15	 & 40.27	 & 36.65	 & 31.45	 & 48.46 \\
    Astec	 & 32.75	 & 36.3	 & 41	 & 36.34	 & 33.33	 & 28.74	 & 46.07 \\
    Parabel	 & 30.21	 & 33.85	 & 38.46	 & 35.26	 & 32.44	 & 28.06	 & 42.84 \\
        \midrule
        \multicolumn{8}{c}{LF-P2PTitles-10M}\\ \midrule							
    \alg	 & \textbf{35.52}	 & \textbf{37.91}	 & \textbf{39.91}	 & \textbf{43.14}	 & \textbf{39.93}	 & \textbf{36.9}	 & \textbf{35.82} \\
    DECAF	 & 20.51	 & 21.38	 & 22.85	 & 28.3	 & 25.75	 & 23.99	 & 20.9 \\
    Astec	 & 20.31	 & 22.16	 & 24.23	 & 29.75	 & 27.49	 & 25.85	 & 22.3 \\
    Parabel	 & 19.99	 & 22.05	 & 24.33	 & 30.22	 & 27.77	 & 26.1	 & 22.81 \\
        \bottomrule
    \end{tabular}}
\end{table}

%% file: tables/knn_graph.tex
\begin{table}
    \caption{An ablation study exploring the benefits of the \metaprop step for other XC methods. Although \alg still provides the leading accuracies, existing methods show consistent gains from the use of the \metaprop step.}
    \label{tab:knngraph}
      \centering
      \resizebox{\linewidth}{!}{
        \begin{tabular}{@{}l|cccc|cccc@{}}
        \toprule
         \textbf{Method} & \textbf{PSP@1}  & \textbf{PSP@5} & \textbf{P@1} &  \textbf{P@5} & \textbf{PSP@1}  & \textbf{PSP@5} & \textbf{P@1} &  \textbf{P@5} \\
         \midrule
\multicolumn{9}{c}{LF-AmazonTitles-131K}\\ \midrule	
& \multicolumn{4}{c}{Original} |& \multicolumn{4}{c}{With GAME}\\ \midrule
\alg & - & - & - & - & \textbf{33.51} & \textbf{44.7} & \textbf{40.74} & \textbf{19.88} \\
Parabel & 23.27 & 32.14 & 32.6 & 15.61 & 24.81	& 34.94 & 33.24	& 16.51\\
AttentionXML & 23.97 & 32.57 & 32.25  & 15.61 & 24.63 &	34.48 & 32.59 & 16.25 \\
\midrule
\multicolumn{9}{c}{LF-WikiSeeAlsoTitles-320K}\\ \midrule	
& \multicolumn{4}{c}{Original} |& \multicolumn{4}{c}{With GAME}\\ \midrule
\alg & - & - & - & - & \textbf{22.01}  & \textbf{26.27} & \textbf{29.35} & \textbf{15.05} \\
Parabel & 9.24 & 11.8 & 17.68 & 8.59 & 10.28 & 13.06 & 17.99	& 9\\
AttentionXML & 9.45  & 11.73 & 17.56  & 8.52 & 10.05 &	12.59 & 17.49 &	8.77 \\
\bottomrule
    \end{tabular}
		}
\end{table}

%% file: figures/fig_tex/results_contrib.tex
\begin{figure}
    \centering
    \includegraphics[width=0.8\linewidth]{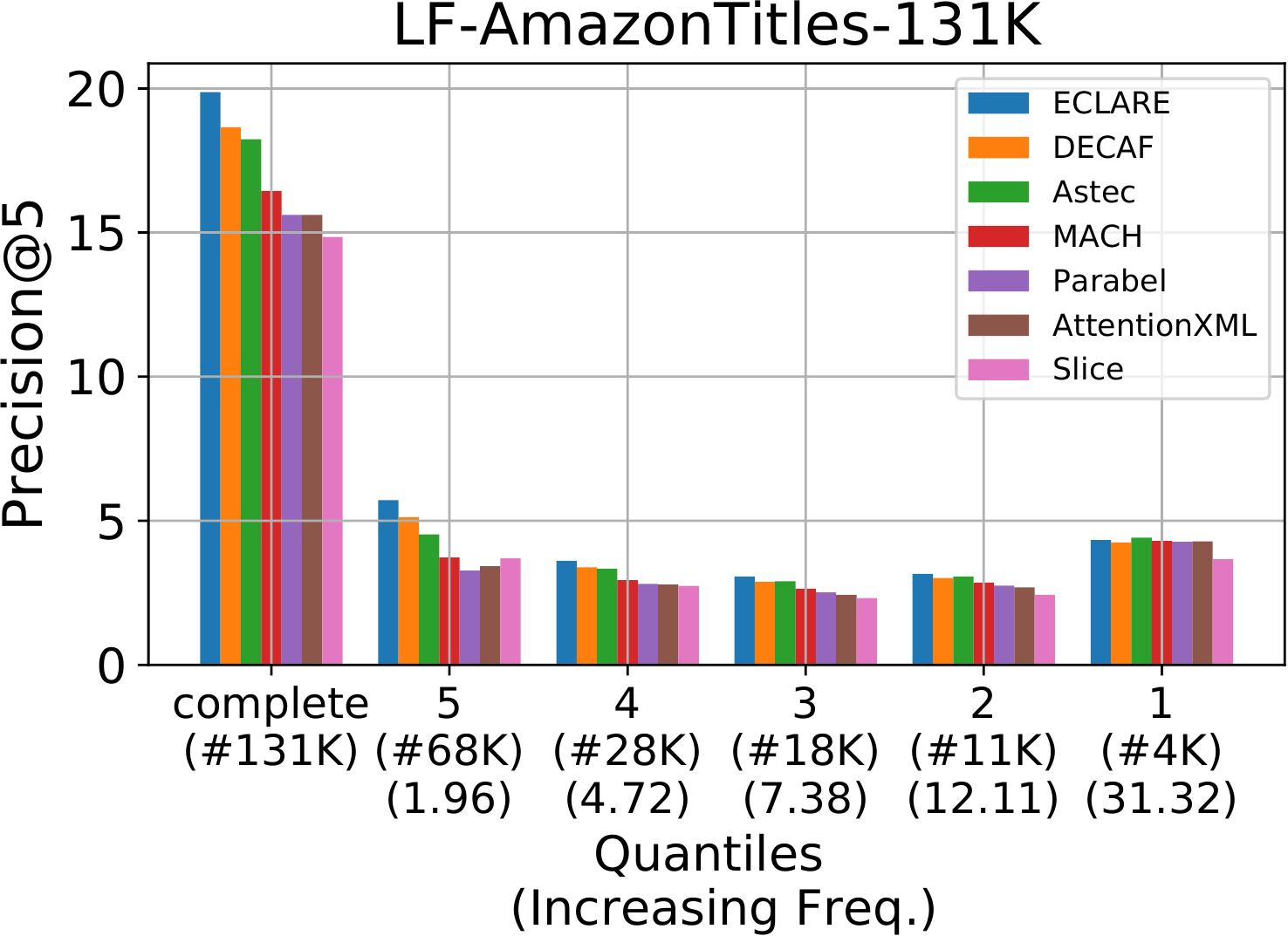}
    \includegraphics[width=0.8\linewidth]{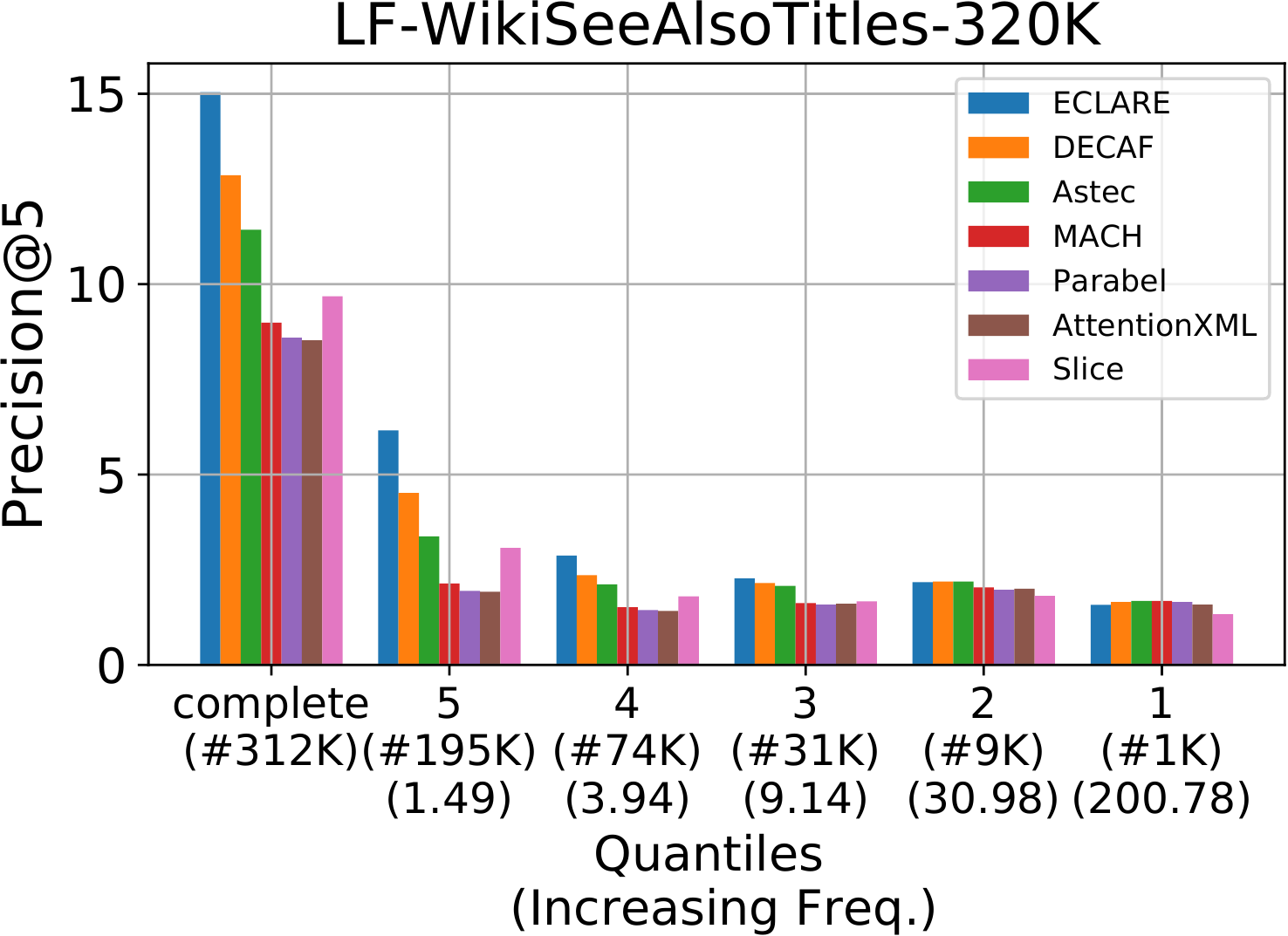}
    \caption{Analyzing the performance of \alg and other methods on popular vs rare labels. Labels were divided into 5 bins in increasing order of popularity. The plots show the overall P@5 for each method (histogram group ``complete'') and how much each bin contributed to this value. \alg clearly draws much of its P@5 performance from the bin with the rarest labels (histogram group 5), i.e \alg's superiority over other methods in terms of P@5 comes from predicting challenging rare labels and not easy-to-predict popular labels. \alg's lead over other methods is also more prominent for rare bins (histogram groups 4, 5).}
    \label{fig:sup:contrib}
\end{figure}

%% file: tables/ablation.tex
\begin{table}
    \caption{An ablation study exploring alternate design decisions. Design choices made by \alg for its components were found to be optimal among popular alternatives.}
    \label{tab:ablation}
      \centering
      \resizebox{\linewidth}{!}{
        \begin{tabular}{@{}l|ccc|ccc@{}}
        \toprule
         \textbf{Method} & \textbf{PSP@1}  & \textbf{PSP@3} & \textbf{PSP@5} & \textbf{P@1}  & \textbf{P@3}  & \textbf{P@5} \\
         \midrule
    \multicolumn{7}{c}{LF-AmazonTitles-131K}\\ \midrule		
\alg	 & \textbf{33.51}	 & \textbf{39.55}	 & \textbf{44.7}	 & \textbf{40.74}	 & \textbf{27.54}	 & \textbf{19.88} \\
\alg-GCN	 & 24.02	 & 29.32	 & 34.02	 & 30.94	 & 21.12	 & 15.52 \\
\alg-LightGCN	 & 31.36	 & 36.79	 & 41.58	 & 38.39	 & 25.59	 & 18.44 \\
\alg-Cooc	 & 32.82	 & 38.67	 & 43.72	 & 39.95	 & 26.9	 & 19.39 \\
\alg-PN	 & 32.49	 & 38.15	 & 43.25	 & 39.63	 & 26.64	 & 19.24 \\
\alg-PPR	 & 12.51	 & 16.42	 & 20.25	 & 14.42	 & 11.04	 & 8.75 \\
\alg-NoGraph	 & 30.49	 & 36.09	 & 41.13	 & 37.45	 & 25.45	 & 18.46 \\
\alg-No\lt	 & 32.3	 & 37.88	 & 42.87	 & 39.33	 & 26.41	 & 19.05 \\
\alg-NoRefine	 & 28.18	 & 33.14	 & 38.3	 & 29.99	 & 21.6	 & 16.32 \\
\alg-SUM	 & 31.45	 & 36.73	 & 41.65	 & 38.02	 & 25.54	 & 18.49 \\
\alg-k=2	 & 32.23	 & 38.06	 & 43.23	 & 39.38	 & 26.57	 & 19.22 \\
\alg-8K	 & 29.98	 & 35.08	 & 39.71	 & 37	 & 24.86	 & 17.93 \\
        \midrule
        \multicolumn{7}{c}{LF-WikiSeeAlsoTItles-320K}\\ \midrule		
    \alg	 & \textbf{22.01}	 & \textbf{24.23}	 & \textbf{26.27}	 & \textbf{29.35}	 & \textbf{19.83}	 & \textbf{15.05} \\
    \alg-GCN	 & 13.76	 & 15.88	 & 17.67	 & 21.76	 & 14.61	 & 11.14 \\
    \alg-LightGCN	 & 19.05	 & 21.24	 & 23.14	 & 26.31	 & 17.64	 & 13.35 \\
    \alg-Cooc	 & 20.96	 & 23.1	 & 25.07	 & 28.54	 & 19.06	 & 14.4 \\
    \alg-PN	 & 20.42	 & 22.56	 & 24.59	 & 28.24	 & 18.88	 & 14.3 \\
    \alg-PPR	 & 4.83	 & 5.53	 & 7.12	 & 5.21	 & 3.82	 & 3.5 \\
    \alg-NoGraph	 & 18.44	 & 20.49	 & 22.42	 & 26.11	 & 17.59	 & 13.35 \\
    \alg-No\lt	 & 20.16	 & 22.22	 & 24.16	 & 27.73	 & 18.48	 & 13.99 \\
    \alg-NoRefine	 & 20.27	 & 21.26	 & 22.8	 & 24.83	 & 16.61	 & 12.66 \\
    \alg-SUM	 & 20.59	 & 22.48	 & 24.36	 & 27.59	 & 18.5	 & 13.99 \\
    \alg-k=2	 & 20.12	 & 22.38	 & 24.43	 & 27.77	 & 18.64	 & 14.14 \\
    \alg-8K	 & 13.42	 & 15.03	 & 16.47	 & 20.31	 & 13.51	 & 10.22 \\
\bottomrule
    \end{tabular}
		}
\end{table}

%% file: figures/fig_tex/clusters.tex
\begin{figure}
    \centering
    \includegraphics[width=\linewidth]{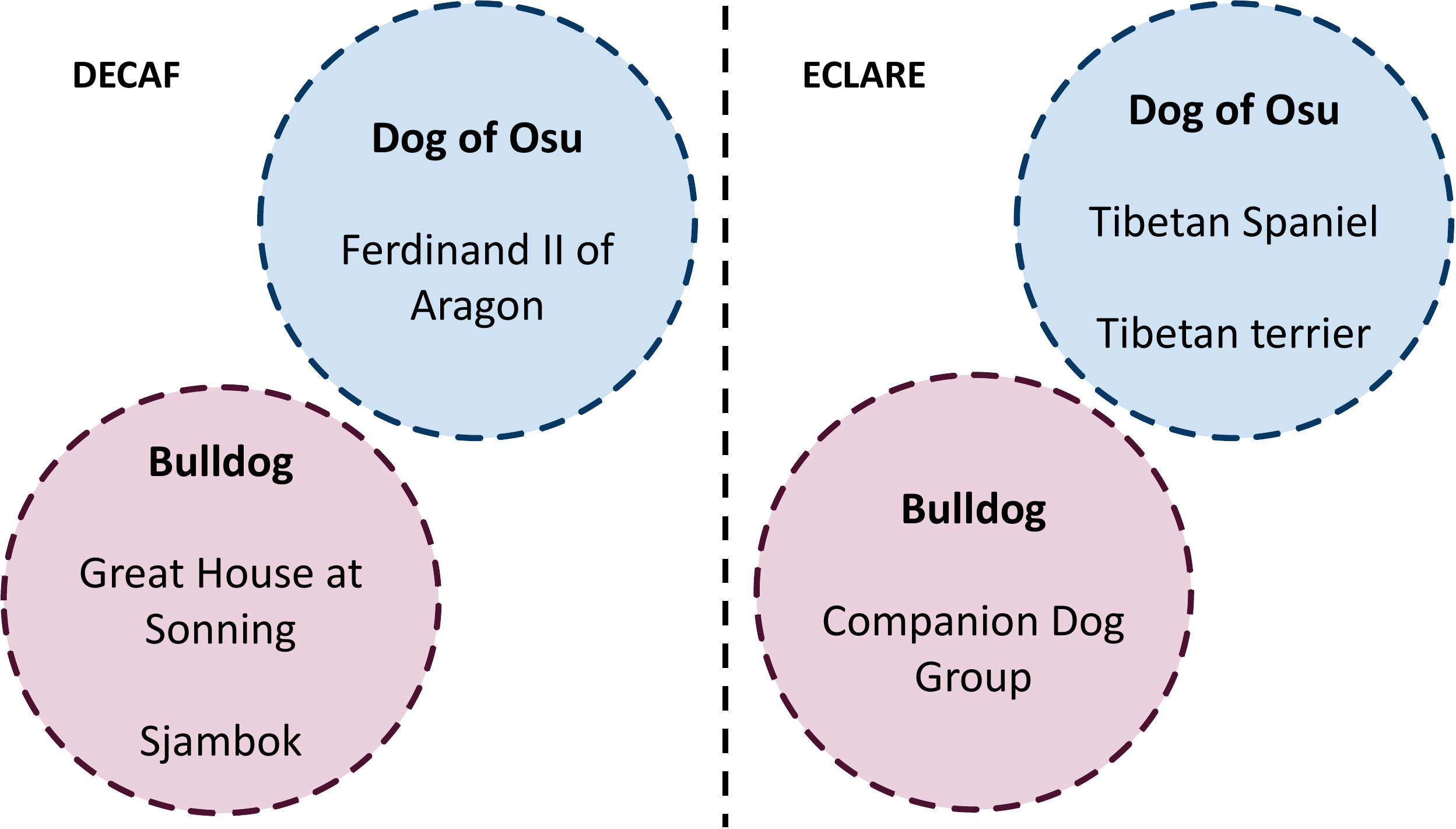}
    \caption{A comparison of meta-label clusters created by \alg compared to those created by DECAF. Note that the clusters for DECAF are rather amorphous, combing labels with diverse intents whereas those for \alg are much more focused. We note that other methods such ASTEC and AttentionXML offered similarly noisy clusters. It is clear that clustering based on label centroids that are augmented using label correlation information creates far superior clusters that do not contain noisy and irrelevant labels.}
    \label{fig:clusters}
\end{figure}

%% file: figures/fig_tex/figure_attention.tex
\begin{figure}
    \centering
    \includegraphics[width=0.9\linewidth]{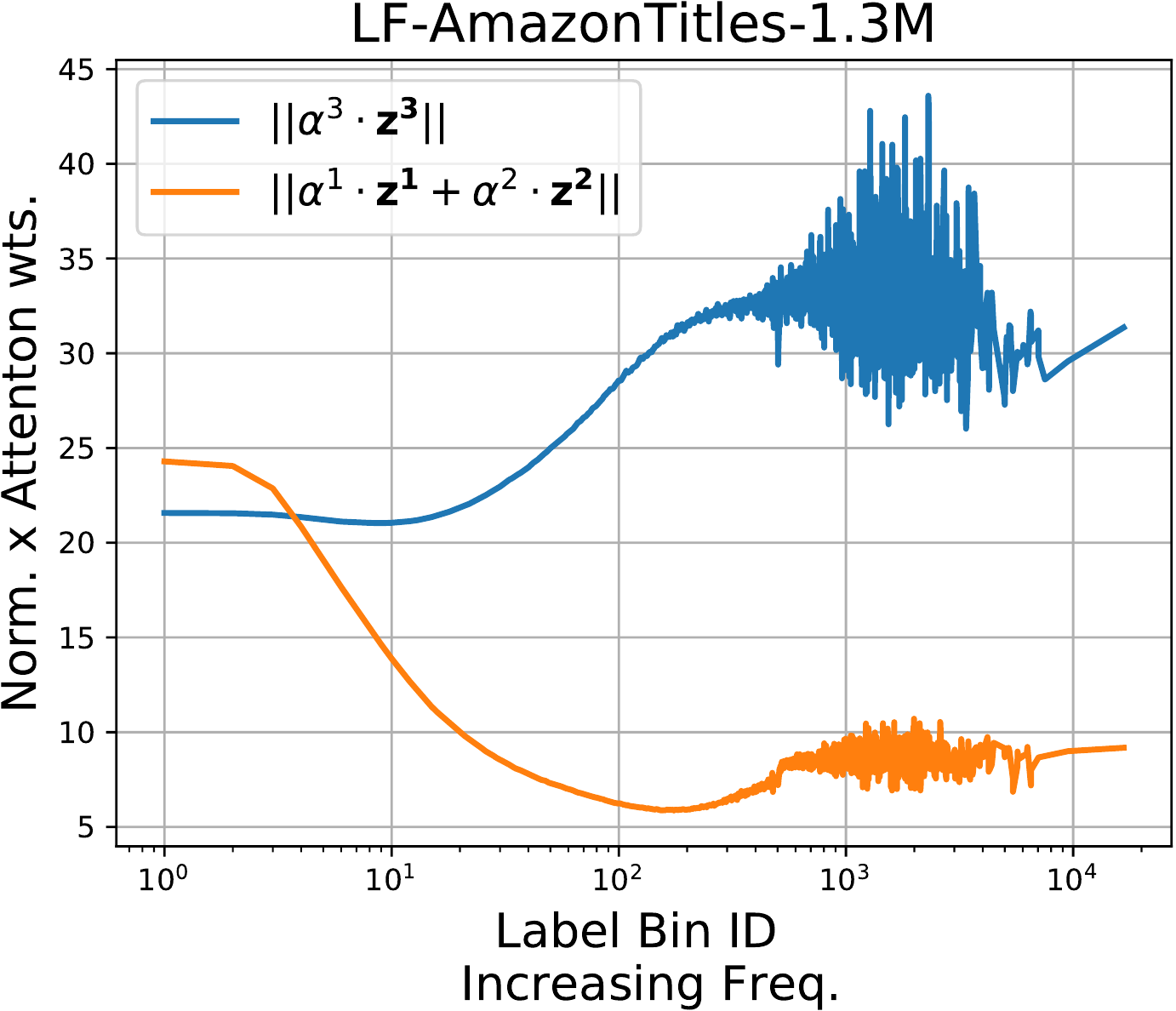}
    \caption{Analysing the dominance of various components in the label classifier $\vw_l$ for rare vs popular labels. For rare labels (on the left), components that focus on label metadata i.e. $\hat\vz^1_l, \hat\vz^2_l$ gain significance whereas for popular labels (on the right), the unrestricted refinement vector $\hat\vz^3_l$ becomes dominant. This illustrates the importance of graph augmentation for data-scarce labels for which the refinement vectors cannot be trained adequately owing to lack of training data.}
    \label{fig:attention}
\end{figure}

%% file: tables/examples.tex
\begin{table*}
    \centering
    \caption{A subjective comparison of the top 5 label predictions by \alg and other algorithms on the WikiSeeAlso-350K and P2PTitles-2M datasets. Predictions typeset in black color were a part of the ground truth whereas those in light gray color were not. \alg is able to offer precise recommendations for extremely rare labels missed by other methods. For instance, the label \notsametext in the first example is so rare that it occurred only twice in the training set. This label does not have any token overlaps with its document or co-occurring labels either. This may have caused techniques such as DECAF that rely solely on label text, to miss such predictions. The examples also establish that incorporating label co-occurrence allows \alg to infer the correct intent of a document or a user query. For instance, in the second example, all other methods, including DECAF, either incorrectly focus on the tokens ``\textsf{Academy Awards}'' in the document title and start predicting labels related to other editions of the Academy Awards, or else amorphous labels about entertainment awards in general. On the other hand, \alg is able to correctly predict other labels corresponding to award ceremonies held in the same year as the 85th Academy awards, as well as the rare label ``\textsf{List of \dots Best Foreign Language Film}''. Similarly, in the third example, \alg correctly determines that the user is interested in faux fur coats and not necessarily in the brand \textsf{Draper's \& Damon's} itself whereas methods such as DECAF that rely solely on label and document text, focus on the brand name alone and start predicting shirts and jackets of the same brand which are irrelevant to the user query.}
    \label{tab:examples}
    \resizebox{\textwidth}{!}{
    \begin{tabular}{p{0.1\textwidth}|p{0.95\textwidth}}
        \toprule
        \textbf{Algorithm} & \textbf{Predictions} \\
        \midrule
        \multicolumn{2}{c}{LF-WikiSeeAlsoTitles-320K} \\
        \midrule
        \midrule
        \textbf{Document} & \textbf{Tibetan Terrier} \\
\alg & Tibetan Spaniel, Tibetan kyi apso, Lhasa Apso, Dog of Osu, Tibetan Mastiff \\
DECAF &  \wpred{Fox Terrier}, Tibetan Spaniel, \wpred{Terrier}, \wpred{Bull Terrier},\wpred{Bulldog}\\

Astec & \wpred{Standard Tibetan}, \wpred{List of domesticated Scottish breeds}, \wpred{List of organizations of Tibetans in exile}, \wpred{Tibet}, \wpred{Riwoche horse}\\
Parabel & \wpred{Tibet}, \wpred{List of organizations of Tibetans in exile}, \wpred{List of domesticated Scottish breeds}, \wpred{History of Tibet}, \wpred{Languages of Bhutan} \\
AttentionXML & \wpred{List of organizations of Tibetans in exile}, \wpred{List of domesticated Scottish breeds},\wpred{Dog}, \wpred{Bull Terrier}, \wpred{Dog crossbreed}
        \\ \bottomrule \toprule
        \textbf{Document} & \textbf{85th Academy Awards} \\
        \midrule
        \alg &  {{List of submissions to the 85th Academy Awards for Best Foreign Language Film}}, 33rd Golden Raspberry Awards, 19th Screen Actors Guild Awards, 67th Tony Awards, 70th Golden Globe Awards \\ 
        DECAF &  \wpred{List of American films of 1956}, \wpred{87th Academy Awards}, \wpred{List of American films of 1957},\wpred{1963 in film}, \wpred{13th Primetime Emmy Awards} \\ 
        Astec & \wpred{65th Tony Awards}, \wpred{29th Primetime Emmy Awards}, \wpred{32nd Golden Raspberry Awards}, \wpred{64th Primetime Emmy Awards}, \wpred{18th Screen Actors Guild Awards}  \\ 
        Parabel & \wpred{1928 in film}, \wpred{1931 in film}, \wpred{1930 in film}, \wpred{48th Academy Awards}, \wpred{26th Primetime Emmy Awards}, \wpred{31st European Film Awards} \\ 
        AttentionXML &  \wpred{29th Primetime Emmy Awards}, \wpred{62nd British Academy Film Awards}, \wpred{60th Primetime Emmy Awards}, \wpred{65th Tony Awards}, \wpred{29th Golden Raspberry Awards}\\
        \bottomrule \toprule
        \multicolumn{2}{c}{P2PTitles-2M} \\
        \midrule
        \midrule
        \textbf{Document} & \textbf{Draper's \& Damon's Women's Chevron Faux Fur Coat Tan L} \\ \midrule
        \alg & Grey Wolf Faux Fur Coat XXL / Grey, Big on Dots Faux-Fur Coat by LUXE, Avec Les Filles Bonded Faux-Fur Long Coat Size Large Black, Roaman's Women's Short Faux-Fur Coat (Black) 1X, \wpred{Dennis Basso Faux Fur Jacket with Stand Collar Size XX-Small Cappuccino}\\ 
        DECAF & \wpred{Draper's \& Damon's Women's Petite Cabana Big Shirt Blue P-L}, \wpred{Draper's \& Damon's Women's Petite Top It Off Stripe Jacket Blue P-L}, \wpred{Draper's \& Damon's Women's Petite Standing Ovation Jacket Black P-L}, \wpred{Draper \& Damon Jackets \& Coats | Draper \& Damons Size L Colorful Coat Wpockets | Color: Black/Green | Size: L}, \wpred{Draper's \& Damon's Women's Impressionist Textured Jacket Multi L}\\ 
        Astec & \wpred{Draper's \& Damon's Women's Petite Cabana Big Shirt Blue P-L}, \wpred{Draper's \& Damon's Women's Petite Top It Off Stripe Jacket Blue P-L}, \wpred{Draper's \& Damon's Women's Impressionist Textured Jacket Multi L}, \wpred{Draper's \& Damon's Women's Embroidered Tulle Jacket Dress Blue 14}, \wpred{Draper's \& Damon's Women's Petite Standing Ovation Jacket Black P-L} \\ 
        Parabel & \wpred{Draper's \& Damon's Women's Impressionist Textured Jacket Multi L}, \wpred{Draper's \& Damon's Women's Over The Rainbow Jacket Multi P-L}, \wpred{Draper's \& Damon's Women's Petite Painted Desert Jacket White P-M}, \wpred{Draper Women's Drapers \& Damons Pants Suit - Pant Suit | Color: Black | Size: L}, \wpred{Draper's \& Damon's Women's Petite Floral \& Stripe Knit Mesh Jacket Scarlet Multi P-L} \\
        \bottomrule
    \end{tabular}
    }
\end{table*}

%% file: tables/LMI.tex
\begin{table}
    \caption{An ablation study showing loss of mutual information (lower is better) using various clustering strategies as well as fanouts. Lowering the number of metalabels $K = \abs\cC$ hurts performance. Competing methods that do not use graph-augmented clustering offer poor LMI, especially MACH that uses random hashes to cluster labels.}
    \label{tab:lmi}
      \centering
      \resizebox{\linewidth}{!}{
        \begin{tabular}{@{}l|cc|cc@{}}
        \toprule
         \textbf{Dataset} & \textbf{\alg}  & \textbf{\alg} & \textbf{DECAF} & \textbf{MACH} \\
         & \textbf{$|\cC|=2^{17}$}  & \textbf{$|\cC|=2^{15}$} & & \\
         \midrule
         LF-AmazonTitles-131K & \textbf{5.44} & 5.82 & 7.40 & 29.68 \\
         LF-WikiSeeAlsoTitles-320K & \textbf{3.96} & 11.31 & 5.47 & 35.31 \\
         \midrule
    \end{tabular}
		}
\end{table}

%% file: sections/conclusion.tex
\section{Conclusion}
\label{sec:conc}
This paper presents the architecture and accompanying training and prediction techniques for the \alg method to perform extreme multi-label classification at the scale of millions of labels. The specific contributions of \alg include a framework for incorporating label graph information at massive scales, as well as critical design and algorithmic choices that enable collaborative learning using label correlation graphs with millions of labels. This includes systematic augmentations to standard XC algorithmic operations such as label-clustering, negative sampling, shortlisting, and re-ranking, to incorporate label correlations in a manner that scales to tasks with millions of labels, all of which were found to be essential to the performance benefits offered by \alg. The creation of label correlation graphs from ground truth data alone and its use in a GCN-style architecture to obtain multiple label representations is critical to \alg's performance benefits. The proposed approach greatly outperforms state-of-the-art XC methods on multiple datasets while still offering millisecond level prediction times even on the largest datasets. Thus, \alg establishes a standard for incorporating label metadata into XC techniques. These findings suggest promising directions for further study including effective graph pruning for heavy tailed datasets, using higher order convolutions ($k > 1$) in a scalable manner, and performing collaborative learning with heterogeneous and even multi-modal label sets. This has the potential to enable generalisation to settings where labels include textual objects such as (related) webpages and documents, but also videos, songs, etc.

%% file: ms.bbl

\begin{thebibliography}{51}


\ifx \showCODEN    \undefined \def \showCODEN     #1{\unskip}     \fi
\ifx \showDOI      \undefined \def \showDOI       #1{#1}\fi
\ifx \showISBNx    \undefined \def \showISBNx     #1{\unskip}     \fi
\ifx \showISBNxiii \undefined \def \showISBNxiii  #1{\unskip}     \fi
\ifx \showISSN     \undefined \def \showISSN      #1{\unskip}     \fi
\ifx \showLCCN     \undefined \def \showLCCN      #1{\unskip}     \fi
\ifx \shownote     \undefined \def \shownote      #1{#1}          \fi
\ifx \showarticletitle \undefined \def \showarticletitle #1{#1}   \fi
\ifx \showURL      \undefined \def \showURL       {\relax}        \fi
\providecommand\bibfield[2]{#2}
\providecommand\bibinfo[2]{#2}
\providecommand\natexlab[1]{#1}
\providecommand\showeprint[2][]{arXiv:#2}

\bibitem[\protect\citeauthoryear{Babbar and Sch\"{o}lkopf}{Babbar and
  Sch\"{o}lkopf}{2017}]%
        {Babbar17}
\bibfield{author}{\bibinfo{person}{R. Babbar} {and} \bibinfo{person}{B.
  Sch\"{o}lkopf}.} \bibinfo{year}{2017}\natexlab{}.
\newblock \showarticletitle{DiSMEC: Distributed Sparse Machines for Extreme
  Multi-label Classification}. In \bibinfo{booktitle}{\emph{WSDM}}.
\newblock


\bibitem[\protect\citeauthoryear{Babbar and Sch\"{o}lkopf}{Babbar and
  Sch\"{o}lkopf}{2019}]%
        {Babbar19}
\bibfield{author}{\bibinfo{person}{R. Babbar} {and} \bibinfo{person}{B.
  Sch\"{o}lkopf}.} \bibinfo{year}{2019}\natexlab{}.
\newblock \showarticletitle{Data scarcity, robustness and extreme multi-label
  classification}.
\newblock \bibinfo{journal}{\emph{ML}} (\bibinfo{year}{2019}).
\newblock


\bibitem[\protect\citeauthoryear{Bhatia, Dahiya, Jain, Mittal, Prabhu, and
  Varma}{Bhatia et~al\mbox{.}}{2016}]%
        {XMLRepo}
\bibfield{author}{\bibinfo{person}{K. Bhatia}, \bibinfo{person}{K. Dahiya},
  \bibinfo{person}{H. Jain}, \bibinfo{person}{A. Mittal}, \bibinfo{person}{Y.
  Prabhu}, {and} \bibinfo{person}{M. Varma}.} \bibinfo{year}{2016}\natexlab{}.
\newblock \bibinfo{title}{The extreme classification repository: Multi-label
  datasets and code}.
\newblock
\newblock
\urldef\tempurl%
\url{http://manikvarma.org/downloads/XC/XMLRepository.html}
\showURL{%
\tempurl}


\bibitem[\protect\citeauthoryear{Bhatia, Jain, Kar, Varma, and Jain}{Bhatia
  et~al\mbox{.}}{2015}]%
        {Bhatia15}
\bibfield{author}{\bibinfo{person}{K. Bhatia}, \bibinfo{person}{H. Jain},
  \bibinfo{person}{P. Kar}, \bibinfo{person}{M. Varma}, {and}
  \bibinfo{person}{P. Jain}.} \bibinfo{year}{2015}\natexlab{}.
\newblock \showarticletitle{Sparse Local Embeddings for Extreme Multi-label
  Classification}. In \bibinfo{booktitle}{\emph{NIPS}}.
\newblock


\bibitem[\protect\citeauthoryear{Bojanowski, Grave, Joulin, and
  Mikolov}{Bojanowski et~al\mbox{.}}{2017}]%
        {Bojanowski17}
\bibfield{author}{\bibinfo{person}{P. Bojanowski}, \bibinfo{person}{E. Grave},
  \bibinfo{person}{A. Joulin}, {and} \bibinfo{person}{T. Mikolov}.}
  \bibinfo{year}{2017}\natexlab{}.
\newblock \showarticletitle{Enriching Word Vectors with Subword Information}.
\newblock \bibinfo{journal}{\emph{Transactions of the Association for
  Computational Linguistics}} (\bibinfo{year}{2017}).
\newblock


\bibitem[\protect\citeauthoryear{Chang, Yu, Zhong, Yang, and Dhillon}{Chang
  et~al\mbox{.}}{2020}]%
        {Chang20}
\bibfield{author}{\bibinfo{person}{W-C. Chang}, \bibinfo{person}{H.-F. Yu},
  \bibinfo{person}{K. Zhong}, \bibinfo{person}{Y. Yang}, {and}
  \bibinfo{person}{I. Dhillon}.} \bibinfo{year}{2020}\natexlab{}.
\newblock \showarticletitle{Taming Pretrained Transformers for Extreme
  Multi-label Text Classification}. In \bibinfo{booktitle}{\emph{KDD}}.
\newblock


\bibitem[\protect\citeauthoryear{Chung}{Chung}{2005}]%
        {chung05}
\bibfield{author}{\bibinfo{person}{F. Chung}.} \bibinfo{year}{2005}\natexlab{}.
\newblock \showarticletitle{Laplacians and the Cheeger inequality for directed
  graphs}.
\newblock \bibinfo{journal}{\emph{Annals of Combinatorics}}
  \bibinfo{volume}{9}, \bibinfo{number}{1} (\bibinfo{year}{2005}),
  \bibinfo{pages}{1--19}.
\newblock


\bibitem[\protect\citeauthoryear{Dahiya, Saini, Mittal, Shaw, Dave, Soni, Jain,
  Agarwal, and Varma}{Dahiya et~al\mbox{.}}{2021}]%
        {Dahiya21}
\bibfield{author}{\bibinfo{person}{K. Dahiya}, \bibinfo{person}{D. Saini},
  \bibinfo{person}{A. Mittal}, \bibinfo{person}{A. Shaw}, \bibinfo{person}{K.
  Dave}, \bibinfo{person}{A. Soni}, \bibinfo{person}{H. Jain},
  \bibinfo{person}{S. Agarwal}, {and} \bibinfo{person}{M. Varma}.}
  \bibinfo{year}{2021}\natexlab{}.
\newblock \showarticletitle{DeepXML: A Deep Extreme Multi-Label Learning
  Framework Applied to Short Text Documents}. In
  \bibinfo{booktitle}{\emph{WSDM}}.
\newblock


\bibitem[\protect\citeauthoryear{Dhillon, Mallela, and Kumar}{Dhillon
  et~al\mbox{.}}{2003}]%
        {dhillon03}
\bibfield{author}{\bibinfo{person}{I.~S. Dhillon}, \bibinfo{person}{S.
  Mallela}, {and} \bibinfo{person}{R. Kumar}.} \bibinfo{year}{2003}\natexlab{}.
\newblock \showarticletitle{{A Divisive Information-Theoretic Feature
  Clustering Algorithm for Text Classification}}.
\newblock \bibinfo{journal}{\emph{JMLR}}  \bibinfo{volume}{3}
  (\bibinfo{year}{2003}), \bibinfo{pages}{1265--1287}.
\newblock


\bibitem[\protect\citeauthoryear{Guo, Mousavi, Wu, Holtmann-Rice, Kale, Reddi,
  and Kumar}{Guo et~al\mbox{.}}{2019}]%
        {Guo2019}
\bibfield{author}{\bibinfo{person}{C. Guo}, \bibinfo{person}{A. Mousavi},
  \bibinfo{person}{X. Wu}, \bibinfo{person}{Daniel~N. Holtmann-Rice},
  \bibinfo{person}{S. Kale}, \bibinfo{person}{S. Reddi}, {and}
  \bibinfo{person}{S. Kumar}.} \bibinfo{year}{2019}\natexlab{}.
\newblock \showarticletitle{Breaking the Glass Ceiling for Embedding-Based
  Classifiers for Large Output Spaces}. In \bibinfo{booktitle}{\emph{Neurips}}.
\newblock


\bibitem[\protect\citeauthoryear{Gupta, Wadbude, Natarajan, Karnick, Jain, and
  Rai}{Gupta et~al\mbox{.}}{2019}]%
        {Gupta19}
\bibfield{author}{\bibinfo{person}{V. Gupta}, \bibinfo{person}{R. Wadbude},
  \bibinfo{person}{N. Natarajan}, \bibinfo{person}{H. Karnick},
  \bibinfo{person}{P. Jain}, {and} \bibinfo{person}{P. Rai}.}
  \bibinfo{year}{2019}\natexlab{}.
\newblock \showarticletitle{Distributional Semantics Meets Multi-Label
  Learning}. In \bibinfo{booktitle}{\emph{AAAI}}.
\newblock


\bibitem[\protect\citeauthoryear{Hamilton, Ying, and Leskovec}{Hamilton
  et~al\mbox{.}}{2017}]%
        {hamilton17}
\bibfield{author}{\bibinfo{person}{W. Hamilton}, \bibinfo{person}{Z. Ying},
  {and} \bibinfo{person}{J. Leskovec}.} \bibinfo{year}{2017}\natexlab{}.
\newblock \showarticletitle{Inductive representation learning on large graphs}.
  In \bibinfo{booktitle}{\emph{NIPS}}. \bibinfo{pages}{1024--1034}.
\newblock


\bibitem[\protect\citeauthoryear{He, Zhang, Ren, and Sun}{He
  et~al\mbox{.}}{2015}]%
        {he2015delving}
\bibfield{author}{\bibinfo{person}{K. He}, \bibinfo{person}{X. Zhang},
  \bibinfo{person}{S. Ren}, {and} \bibinfo{person}{J. Sun}.}
  \bibinfo{year}{2015}\natexlab{}.
\newblock \showarticletitle{{Delving deep into rectifiers: Surpassing
  human-level performance on imagenet classification}}. In
  \bibinfo{booktitle}{\emph{Proceedings of the IEEE international conference on
  computer vision}}. \bibinfo{pages}{1026--1034}.
\newblock


\bibitem[\protect\citeauthoryear{He, Deng, Wang, Li, Zhang, and Wang}{He
  et~al\mbox{.}}{2020}]%
        {He2020}
\bibfield{author}{\bibinfo{person}{X. He}, \bibinfo{person}{K. Deng},
  \bibinfo{person}{X. Wang}, \bibinfo{person}{Y. Li}, \bibinfo{person}{Y.
  Zhang}, {and} \bibinfo{person}{M. Wang}.} \bibinfo{year}{2020}\natexlab{}.
\newblock \showarticletitle{LightGCN: Simplifying and Powering Graph
  Convolution Network for Recommendation}. In
  \bibinfo{booktitle}{\emph{Proceedings of the 43rd International ACM SIGIR
  Conference on Research and Development in Information Retrieval}}
  \emph{(\bibinfo{series}{SIGIR '20})}.
\newblock


\bibitem[\protect\citeauthoryear{Jain, Balasubramanian, Chunduri, and
  Varma}{Jain et~al\mbox{.}}{2019}]%
        {Jain19}
\bibfield{author}{\bibinfo{person}{H. Jain}, \bibinfo{person}{V.
  Balasubramanian}, \bibinfo{person}{B. Chunduri}, {and} \bibinfo{person}{M.
  Varma}.} \bibinfo{year}{2019}\natexlab{}.
\newblock \showarticletitle{Slice: Scalable Linear Extreme Classifiers trained
  on 100 Million Labels for Related Searches}. In
  \bibinfo{booktitle}{\emph{WSDM}}.
\newblock


\bibitem[\protect\citeauthoryear{Jain, Prabhu, and Varma}{Jain
  et~al\mbox{.}}{2016}]%
        {Jain16}
\bibfield{author}{\bibinfo{person}{H. Jain}, \bibinfo{person}{Y. Prabhu}, {and}
  \bibinfo{person}{M. Varma}.} \bibinfo{year}{2016}\natexlab{}.
\newblock \showarticletitle{Extreme Multi-label Loss Functions for
  Recommendation, Tagging, Ranking and Other Missing Label Applications}. In
  \bibinfo{booktitle}{\emph{KDD}}.
\newblock


\bibitem[\protect\citeauthoryear{Jain, Modhe, and Rai}{Jain
  et~al\mbox{.}}{2017}]%
        {Jain17}
\bibfield{author}{\bibinfo{person}{V. Jain}, \bibinfo{person}{N. Modhe}, {and}
  \bibinfo{person}{P. Rai}.} \bibinfo{year}{2017}\natexlab{}.
\newblock \showarticletitle{Scalable Generative Models for Multi-label Learning
  with Missing Labels}. In \bibinfo{booktitle}{\emph{ICML}}.
\newblock


\bibitem[\protect\citeauthoryear{Jalan and Kar}{Jalan and Kar}{2019}]%
        {Jalan2019}
\bibfield{author}{\bibinfo{person}{A. Jalan} {and} \bibinfo{person}{P. Kar}.}
  \bibinfo{year}{2019}\natexlab{}.
\newblock \showarticletitle{Accelerating Extreme Classification via Adaptive
  Feature Agglomeration}.
\newblock \bibinfo{journal}{\emph{IJCAI}} (\bibinfo{year}{2019}).
\newblock


\bibitem[\protect\citeauthoryear{Jasinska, Dembczynski, Busa-Fekete,
  Pfannschmidt, Klerx, and Hullermeier}{Jasinska et~al\mbox{.}}{2016}]%
        {Jasinska16}
\bibfield{author}{\bibinfo{person}{K. Jasinska}, \bibinfo{person}{K.
  Dembczynski}, \bibinfo{person}{R. Busa-Fekete}, \bibinfo{person}{K.
  Pfannschmidt}, \bibinfo{person}{T. Klerx}, {and} \bibinfo{person}{E.
  Hullermeier}.} \bibinfo{year}{2016}\natexlab{}.
\newblock \showarticletitle{Extreme F-measure Maximization using Sparse
  Probability Estimates}. In \bibinfo{booktitle}{\emph{ICML}}.
\newblock


\bibitem[\protect\citeauthoryear{Joulin, Grave, Bojanowski, and Mikolov}{Joulin
  et~al\mbox{.}}{2017}]%
        {Joulin17}
\bibfield{author}{\bibinfo{person}{A. Joulin}, \bibinfo{person}{E. Grave},
  \bibinfo{person}{P. Bojanowski}, {and} \bibinfo{person}{T. Mikolov}.}
  \bibinfo{year}{2017}\natexlab{}.
\newblock \showarticletitle{Bag of Tricks for Efficient Text Classification}.
  In \bibinfo{booktitle}{\emph{Proceedings of the European Chapter of the
  Association for Computational Linguistics}}.
\newblock


\bibitem[\protect\citeauthoryear{Kanagal, Ahmed, Pandey, Josifovski, Yuan, and
  Garcia-Pueyo}{Kanagal et~al\mbox{.}}{2012}]%
        {Kanagal12}
\bibfield{author}{\bibinfo{person}{Bhargav Kanagal}, \bibinfo{person}{Amr
  Ahmed}, \bibinfo{person}{Sandeep Pandey}, \bibinfo{person}{Vanja Josifovski},
  \bibinfo{person}{Jeff Yuan}, {and} \bibinfo{person}{Lluis Garcia-Pueyo}.}
  \bibinfo{year}{2012}\natexlab{}.
\newblock \showarticletitle{Supercharging Recommender Systems Using Taxonomies
  for Learning User Purchase Behavior}.
\newblock \bibinfo{journal}{\emph{VLDB}} (\bibinfo{date}{June}
  \bibinfo{year}{2012}).
\newblock


\bibitem[\protect\citeauthoryear{Khandagale, Xiao, and Babbar}{Khandagale
  et~al\mbox{.}}{2019}]%
        {Khandagale19}
\bibfield{author}{\bibinfo{person}{S. Khandagale}, \bibinfo{person}{H. Xiao},
  {and} \bibinfo{person}{R. Babbar}.} \bibinfo{year}{2019}\natexlab{}.
\newblock \showarticletitle{Bonsai - Diverse and Shallow Trees for Extreme
  Multi-label Classification}.
\newblock \bibinfo{journal}{\emph{CoRR}} (\bibinfo{year}{2019}).
\newblock


\bibitem[\protect\citeauthoryear{Kingma and Ba}{Kingma and Ba}{2014}]%
        {Kingma14}
\bibfield{author}{\bibinfo{person}{P.~D. Kingma} {and} \bibinfo{person}{J.
  Ba}.} \bibinfo{year}{2014}\natexlab{}.
\newblock \showarticletitle{Adam: {A} Method for Stochastic Optimization}.
\newblock \bibinfo{journal}{\emph{CoRR}} (\bibinfo{year}{2014}).
\newblock


\bibitem[\protect\citeauthoryear{Kipf and Welling}{Kipf and Welling}{2017}]%
        {Kipf17}
\bibfield{author}{\bibinfo{person}{T.~N. Kipf} {and} \bibinfo{person}{M.
  Welling}.} \bibinfo{year}{2017}\natexlab{}.
\newblock \showarticletitle{Semi-Supervised Classification with Graph
  Convolutional Networks}. In \bibinfo{booktitle}{\emph{5th International
  Conference on Learning Representations, {ICLR} 2017, Toulon, France, April
  24-26, 2017, Conference Track Proceedings}}.
\newblock


\bibitem[\protect\citeauthoryear{Klicpera, Bojchevski, and
  G{\"u}nnemann}{Klicpera et~al\mbox{.}}{2018}]%
        {klicpera18}
\bibfield{author}{\bibinfo{person}{J. Klicpera}, \bibinfo{person}{A.
  Bojchevski}, {and} \bibinfo{person}{S. G{\"u}nnemann}.}
  \bibinfo{year}{2018}\natexlab{}.
\newblock \showarticletitle{Predict then Propagate: Graph Neural Networks meet
  Personalized PageRank}. In \bibinfo{booktitle}{\emph{International Conference
  on Learning Representations}}.
\newblock


\bibitem[\protect\citeauthoryear{Liu, Chang, Wu, and Yang}{Liu
  et~al\mbox{.}}{2017}]%
        {Liu17}
\bibfield{author}{\bibinfo{person}{J. Liu}, \bibinfo{person}{W. Chang},
  \bibinfo{person}{Y. Wu}, {and} \bibinfo{person}{Y. Yang}.}
  \bibinfo{year}{2017}\natexlab{}.
\newblock \showarticletitle{Deep Learning for Extreme Multi-label Text
  Classification}. In \bibinfo{booktitle}{\emph{SIGIR}}.
\newblock


\bibitem[\protect\citeauthoryear{Liu, Ott, Goyal, Du, Joshi, Chen, Levy, Lewis,
  Zettlemoyer, and Stoyanov}{Liu et~al\mbox{.}}{2019}]%
        {roberta}
\bibfield{author}{\bibinfo{person}{Yinhan Liu}, \bibinfo{person}{Myle Ott},
  \bibinfo{person}{Naman Goyal}, \bibinfo{person}{Jingfei Du},
  \bibinfo{person}{Mandar Joshi}, \bibinfo{person}{Danqi Chen},
  \bibinfo{person}{Omer Levy}, \bibinfo{person}{Mike Lewis},
  \bibinfo{person}{Luke Zettlemoyer}, {and} \bibinfo{person}{Veselin
  Stoyanov}.} \bibinfo{year}{2019}\natexlab{}.
\newblock \bibinfo{title}{{RoBERTa: A Robustly Optimized BERT Pretraining
  Approach}}.
\newblock \bibinfo{howpublished}{arXiv:1907.11692}.
\newblock


\bibitem[\protect\citeauthoryear{Medini, Huang, Wang, Mohan, and
  Shrivastava}{Medini et~al\mbox{.}}{2019}]%
        {Medini2019}
\bibfield{author}{\bibinfo{person}{T.~K.~R. Medini}, \bibinfo{person}{Q.
  Huang}, \bibinfo{person}{Y. Wang}, \bibinfo{person}{V. Mohan}, {and}
  \bibinfo{person}{A. Shrivastava}.} \bibinfo{year}{2019}\natexlab{}.
\newblock \showarticletitle{Extreme Classification in Log Memory using
  Count-Min Sketch: A Case Study of Amazon Search with 50M Products}. In
  \bibinfo{booktitle}{\emph{Neurips}}.
\newblock


\bibitem[\protect\citeauthoryear{Menon, Chitrapura, Garg, Agarwal, and
  Kota}{Menon et~al\mbox{.}}{2011}]%
        {menon11}
\bibfield{author}{\bibinfo{person}{Aditya~Krishna Menon},
  \bibinfo{person}{Krishna-Prasad Chitrapura}, \bibinfo{person}{Sachin Garg},
  \bibinfo{person}{Deepak Agarwal}, {and} \bibinfo{person}{Nagaraj Kota}.}
  \bibinfo{year}{2011}\natexlab{}.
\newblock \showarticletitle{Response Prediction Using Collaborative Filtering
  with Hierarchies and Side-Information}. In \bibinfo{booktitle}{\emph{KDD}}.
\newblock


\bibitem[\protect\citeauthoryear{Mikolov, Sutskever, Chen, Corrado, and
  Dean}{Mikolov et~al\mbox{.}}{2013}]%
        {Mikolov13}
\bibfield{author}{\bibinfo{person}{T. Mikolov}, \bibinfo{person}{I. Sutskever},
  \bibinfo{person}{K. Chen}, \bibinfo{person}{G. Corrado}, {and}
  \bibinfo{person}{J. Dean}.} \bibinfo{year}{2013}\natexlab{}.
\newblock \showarticletitle{Distributed Representations of Words and Phrases
  and Their Compositionality}. In \bibinfo{booktitle}{\emph{NIPS}}.
\newblock


\bibitem[\protect\citeauthoryear{Mittal, Dahiya, Agrawal, Saini, Agarwal, Kar,
  and Varma}{Mittal et~al\mbox{.}}{2021}]%
        {mittal20}
\bibfield{author}{\bibinfo{person}{A. Mittal}, \bibinfo{person}{K. Dahiya},
  \bibinfo{person}{S. Agrawal}, \bibinfo{person}{D. Saini}, \bibinfo{person}{S.
  Agarwal}, \bibinfo{person}{P. Kar}, {and} \bibinfo{person}{M. Varma}.}
  \bibinfo{year}{2021}\natexlab{}.
\newblock \showarticletitle{{DECAF: Deep Extreme Classification with Label
  Features}}. In \bibinfo{booktitle}{\emph{WSDM}}.
\newblock


\bibitem[\protect\citeauthoryear{Miyato, Kataoka, Koyama, and Yoshida}{Miyato
  et~al\mbox{.}}{2018}]%
        {Miyato18}
\bibfield{author}{\bibinfo{person}{T. Miyato}, \bibinfo{person}{T. Kataoka},
  \bibinfo{person}{M. Koyama}, {and} \bibinfo{person}{Y. Yoshida}.}
  \bibinfo{year}{2018}\natexlab{}.
\newblock \showarticletitle{Spectral Normalization for Generative Adversarial
  Networks}. In \bibinfo{booktitle}{\emph{ICLR}}.
\newblock


\bibitem[\protect\citeauthoryear{Niculescu-Mizil and
  Abbasnejad}{Niculescu-Mizil and Abbasnejad}{2017}]%
        {Niculescu17}
\bibfield{author}{\bibinfo{person}{A. Niculescu-Mizil} {and}
  \bibinfo{person}{E. Abbasnejad}.} \bibinfo{year}{2017}\natexlab{}.
\newblock \showarticletitle{Label Filters for Large Scale Multilabel
  Classification}. In \bibinfo{booktitle}{\emph{AISTATS}}.
\newblock


\bibitem[\protect\citeauthoryear{Pal, Eksombatchai, Zhou, Zhao, Rosenberg, and
  Leskovec}{Pal et~al\mbox{.}}{2020}]%
        {pal20}
\bibfield{author}{\bibinfo{person}{A. Pal}, \bibinfo{person}{C. Eksombatchai},
  \bibinfo{person}{Y. Zhou}, \bibinfo{person}{B. Zhao}, \bibinfo{person}{C.
  Rosenberg}, {and} \bibinfo{person}{J. Leskovec}.}
  \bibinfo{year}{2020}\natexlab{}.
\newblock \showarticletitle{PinnerSage: Multi-Modal User Embedding Framework
  for Recommendations at Pinterest}. In \bibinfo{booktitle}{\emph{KDD '20}}
  (Virtual Event, CA, USA) \emph{(\bibinfo{series}{KDD '20})}.
  \bibinfo{publisher}{Association for Computing Machinery},
  \bibinfo{address}{New York, NY, USA}, \bibinfo{pages}{2311–2320}.
\newblock
\showISBNx{9781450379984}
\urldef\tempurl%
\url{https://doi.org/10.1145/3394486.3403280}
\showDOI{\tempurl}


\bibitem[\protect\citeauthoryear{Prabhu, Kag, Gopinath, Dahiya, Harsola,
  Agrawal, and Varma}{Prabhu et~al\mbox{.}}{2018a}]%
        {Prabhu18}
\bibfield{author}{\bibinfo{person}{Y. Prabhu}, \bibinfo{person}{A. Kag},
  \bibinfo{person}{S. Gopinath}, \bibinfo{person}{K. Dahiya},
  \bibinfo{person}{S. Harsola}, \bibinfo{person}{R. Agrawal}, {and}
  \bibinfo{person}{M. Varma}.} \bibinfo{year}{2018}\natexlab{a}.
\newblock \showarticletitle{Extreme multi-label learning with label features
  for warm-start tagging, ranking and recommendation}. In
  \bibinfo{booktitle}{\emph{WSDM}}.
\newblock


\bibitem[\protect\citeauthoryear{Prabhu, Kag, Harsola, Agrawal, and
  Varma}{Prabhu et~al\mbox{.}}{2018b}]%
        {Prabhu18b}
\bibfield{author}{\bibinfo{person}{Y. Prabhu}, \bibinfo{person}{A. Kag},
  \bibinfo{person}{S. Harsola}, \bibinfo{person}{R. Agrawal}, {and}
  \bibinfo{person}{M. Varma}.} \bibinfo{year}{2018}\natexlab{b}.
\newblock \showarticletitle{Parabel: Partitioned label trees for extreme
  classification with application to dynamic search advertising}. In
  \bibinfo{booktitle}{\emph{WWW}}.
\newblock


\bibitem[\protect\citeauthoryear{Prabhu and Varma}{Prabhu and Varma}{2014}]%
        {Prabhu14}
\bibfield{author}{\bibinfo{person}{Y. Prabhu} {and} \bibinfo{person}{M.
  Varma}.} \bibinfo{year}{2014}\natexlab{}.
\newblock \showarticletitle{FastXML: A Fast, Accurate and Stable
  Tree-classifier for eXtreme Multi-label Learning}. In
  \bibinfo{booktitle}{\emph{KDD}}.
\newblock


\bibitem[\protect\citeauthoryear{Sachdeva, Gupta, and Pudi}{Sachdeva
  et~al\mbox{.}}{2018}]%
        {sachdeva19}
\bibfield{author}{\bibinfo{person}{Noveen Sachdeva}, \bibinfo{person}{Kartik
  Gupta}, {and} \bibinfo{person}{Vikram Pudi}.}
  \bibinfo{year}{2018}\natexlab{}.
\newblock \showarticletitle{Attentive Neural Architecture Incorporating Song
  Features for Music Recommendation}. In \bibinfo{booktitle}{\emph{RecSys}}.
\newblock


\bibitem[\protect\citeauthoryear{Schuster and Nakajima}{Schuster and
  Nakajima}{2012}]%
        {schuster12}
\bibfield{author}{\bibinfo{person}{M. Schuster} {and} \bibinfo{person}{K.
  Nakajima}.} \bibinfo{year}{2012}\natexlab{}.
\newblock \showarticletitle{Japanese and korean voice search}. In
  \bibinfo{booktitle}{\emph{2012 IEEE International Conference on Acoustics,
  Speech and Signal Processing (ICASSP)}}. IEEE, \bibinfo{pages}{5149--5152}.
\newblock


\bibitem[\protect\citeauthoryear{Siblini, Kuntz, and Meyer}{Siblini
  et~al\mbox{.}}{2018}]%
        {Siblini18a}
\bibfield{author}{\bibinfo{person}{W. Siblini}, \bibinfo{person}{P. Kuntz},
  {and} \bibinfo{person}{F. Meyer}.} \bibinfo{year}{2018}\natexlab{}.
\newblock \showarticletitle{CRAFTML, an Efficient Clustering-based Random
  Forest for Extreme Multi-label Learning}. In
  \bibinfo{booktitle}{\emph{ICML}}.
\newblock


\bibitem[\protect\citeauthoryear{Tagami}{Tagami}{2017}]%
        {Tagami17}
\bibfield{author}{\bibinfo{person}{Y. Tagami}.}
  \bibinfo{year}{2017}\natexlab{}.
\newblock \showarticletitle{AnnexML: Approximate Nearest Neighbor Search for
  Extreme Multi-label Classification}. In \bibinfo{booktitle}{\emph{KDD}}.
\newblock


\bibitem[\protect\citeauthoryear{Tang, Jiang, Xia, Pitera, J., and Chawla}{Tang
  et~al\mbox{.}}{2020}]%
        {Tang20}
\bibfield{author}{\bibinfo{person}{P. Tang}, \bibinfo{person}{M. Jiang},
  \bibinfo{person}{B. Xia}, \bibinfo{person}{J.~W. Pitera},
  \bibinfo{person}{Welser J.}, {and} \bibinfo{person}{N.~V. Chawla}.}
  \bibinfo{year}{2020}\natexlab{}.
\newblock \showarticletitle{Multi-Label Patent Categorization with Non-Local
  Attention-Based Graph Convolutional Network}. In
  \bibinfo{booktitle}{\emph{AAAI, 2020}}.
\newblock


\bibitem[\protect\citeauthoryear{Veli{\'{c}}kovi{\'{c}}, Cucurull, Casanova,
  Romero, Li{\`{o}}, and Bengio}{Veli{\'{c}}kovi{\'{c}} et~al\mbox{.}}{2018}]%
        {velickovic18}
\bibfield{author}{\bibinfo{person}{P. Veli{\'{c}}kovi{\'{c}}},
  \bibinfo{person}{G. Cucurull}, \bibinfo{person}{A. Casanova},
  \bibinfo{person}{A. Romero}, \bibinfo{person}{P. Li{\`{o}}}, {and}
  \bibinfo{person}{Y. Bengio}.} \bibinfo{year}{2018}\natexlab{}.
\newblock \showarticletitle{{Graph Attention Networks}}.
\newblock \bibinfo{journal}{\emph{ICLR}} (\bibinfo{year}{2018}).
\newblock


\bibitem[\protect\citeauthoryear{Wydmuch, Jasinska, Kuznetsov, Busa-Fekete, and
  Dembczynski}{Wydmuch et~al\mbox{.}}{2018}]%
        {Wydmuch18}
\bibfield{author}{\bibinfo{person}{M. Wydmuch}, \bibinfo{person}{K. Jasinska},
  \bibinfo{person}{M. Kuznetsov}, \bibinfo{person}{R. Busa-Fekete}, {and}
  \bibinfo{person}{K. Dembczynski}.} \bibinfo{year}{2018}\natexlab{}.
\newblock \showarticletitle{A no-regret generalization of hierarchical softmax
  to extreme multi-label classification}. In \bibinfo{booktitle}{\emph{NIPS}}.
\newblock


\bibitem[\protect\citeauthoryear{Yen, Huang, Dai, Ravikumar, and Xing}{Yen
  et~al\mbox{.}}{2017}]%
        {Yen17}
\bibfield{author}{\bibinfo{person}{E.H.~I. Yen}, \bibinfo{person}{X. Huang},
  \bibinfo{person}{W. Dai}, \bibinfo{person}{I. Ravikumar, P.and~Dhillon},
  {and} \bibinfo{person}{E. Xing}.} \bibinfo{year}{2017}\natexlab{}.
\newblock \showarticletitle{PPDSparse: A Parallel Primal-Dual Sparse Method for
  Extreme Classification}. In \bibinfo{booktitle}{\emph{KDD}}.
\newblock


\bibitem[\protect\citeauthoryear{Yen, Huang, Zhong, Ravikumar, and Dhillon}{Yen
  et~al\mbox{.}}{2016}]%
        {Yen16}
\bibfield{author}{\bibinfo{person}{E.H.~I. Yen}, \bibinfo{person}{X. Huang},
  \bibinfo{person}{K. Zhong}, \bibinfo{person}{P. Ravikumar}, {and}
  \bibinfo{person}{I.~S. Dhillon}.} \bibinfo{year}{2016}\natexlab{}.
\newblock \showarticletitle{PD-Sparse: A Primal and Dual Sparse Approach to
  Extreme Multiclass and Multilabel Classification}. In
  \bibinfo{booktitle}{\emph{ICML}}.
\newblock


\bibitem[\protect\citeauthoryear{Yen, Kale, Yu, Holtmann~R., Kumar, and
  Ravikumar}{Yen et~al\mbox{.}}{2018}]%
        {Yen18a}
\bibfield{author}{\bibinfo{person}{I. Yen}, \bibinfo{person}{S. Kale},
  \bibinfo{person}{F. Yu}, \bibinfo{person}{D. Holtmann~R.},
  \bibinfo{person}{S. Kumar}, {and} \bibinfo{person}{P. Ravikumar}.}
  \bibinfo{year}{2018}\natexlab{}.
\newblock \showarticletitle{Loss Decomposition for Fast Learning in Large
  Output Spaces}. In \bibinfo{booktitle}{\emph{ICML}}.
\newblock


\bibitem[\protect\citeauthoryear{Ying, He, Chen, Eksombatchai, Hamilton, and
  Leskovec}{Ying et~al\mbox{.}}{2018}]%
        {ying2018}
\bibfield{author}{\bibinfo{person}{R. Ying}, \bibinfo{person}{R. He},
  \bibinfo{person}{K. Chen}, \bibinfo{person}{P. Eksombatchai},
  \bibinfo{person}{W. Hamilton}, {and} \bibinfo{person}{J. Leskovec}.}
  \bibinfo{year}{2018}\natexlab{}.
\newblock \showarticletitle{Graph convolutional neural networks for web-scale
  recommender systems}. In \bibinfo{booktitle}{\emph{Proceedings of the 24th
  ACM SIGKDD International Conference on Knowledge Discovery \& Data Mining}}.
  \bibinfo{pages}{974--983}.
\newblock


\bibitem[\protect\citeauthoryear{You, Dai, Zhang, Mamitsuka, and Zhu}{You
  et~al\mbox{.}}{2018}]%
        {You18}
\bibfield{author}{\bibinfo{person}{R. You}, \bibinfo{person}{S. Dai},
  \bibinfo{person}{Z. Zhang}, \bibinfo{person}{H. Mamitsuka}, {and}
  \bibinfo{person}{S. Zhu}.} \bibinfo{year}{2018}\natexlab{}.
\newblock \showarticletitle{AttentionXML: Extreme Multi-Label Text
  Classification with Multi-Label Attention Based Recurrent Neural Networks}.
\newblock \bibinfo{journal}{\emph{CoRR}} (\bibinfo{year}{2018}).
\newblock


\bibitem[\protect\citeauthoryear{Yu, Jain, Kar, and Dhillon}{Yu
  et~al\mbox{.}}{2014}]%
        {Yu14}
\bibfield{author}{\bibinfo{person}{H. Yu}, \bibinfo{person}{P. Jain},
  \bibinfo{person}{P. Kar}, {and} \bibinfo{person}{I.~S. Dhillon}.}
  \bibinfo{year}{2014}\natexlab{}.
\newblock \showarticletitle{Large-scale Multi-label Learning with Missing
  Labels}. In \bibinfo{booktitle}{\emph{ICML}}.
\newblock


\bibitem[\protect\citeauthoryear{Zhang, Cui, and Zhu}{Zhang
  et~al\mbox{.}}{2020}]%
        {zhang20}
\bibfield{author}{\bibinfo{person}{Z. Zhang}, \bibinfo{person}{P. Cui}, {and}
  \bibinfo{person}{W. Zhu}.} \bibinfo{year}{2020}\natexlab{}.
\newblock \showarticletitle{Deep learning on graphs: A survey}.
\newblock \bibinfo{journal}{\emph{IEEE Transactions on Knowledge and Data
  Engineering}} (\bibinfo{year}{2020}).
\newblock


\end{thebibliography}
